%% file: main.tex
\documentclass[sn-mathphys]{sn-jnl}
\jyear{2021}%

\theoremstyle{thmstyleone}%
\theoremstyle{thmstyletwo}%

\theoremstyle{thmstylethree}%

\usepackage{dirtytalk}
\usepackage{caption}
\usepackage{subcaption}
\usepackage{numprint}
\npthousandsep{,}
\npproductsign{\times}
\usepackage{textgreek}
\usepackage{makecell}
\usepackage{tabularx,ragged2e}
\usepackage{multirow}
\usepackage{booktabs}
\usepackage{physics}
\newcolumntype{C}{>{\Centering\arraybackslash}X}  
\newcolumntype{s}{>{\Centering\hsize=.5\hsize}X}
\newcolumntype{k}{>{\Centering\hsize=.26\hsize}X}
\newcolumntype{j}{>{\Centering\hsize=.2\hsize}X}
\newcolumntype{y}{>{\Centering\hsize=.14\hsize}X}
\newcolumntype{t}{>{\Centering\hsize=.07\hsize}X}
\newcolumntype{i}{>{\Centering\hsize=.05\hsize}X}
\newcommand{\fnc}[1]{\operatorname{#1}}
\usepackage{textcomp}
\usepackage{mathtools,bm,amssymb}

\usepackage{lineno}

\begin{document}

\title[Medical records condensation: a roadmap towards healthcare data democratisation]{Medical records condensation: a roadmap towards healthcare data democratisation}

\author[1,4]{\fnm{Yujiang} \sur{Wang}}\email{yujiang.wang@oxford-oscar.cn}

\author*[1]{\fnm{Anshul} \sur{Thakur}}\email{anshul.thakur@eng.ox.ac.uk}

\author[2]{\fnm{Mingzhi} \sur{Dong}}\email{mingzhidong@gmail.com} 

\author[3]{\fnm{Pingchuan} \sur{Ma}}\email{pingchuan.ma16@imperial.ac.uk}

\author[3]{\fnm{Stavros} \sur{Petridis}}\email{stavros.petridis04@imperial.ac.uk}

\author*[2]{\fnm{Li} \sur{Shang}}\email{lishang@fudan.edu.cn}

\author[1]{\fnm{Tingting} \sur{Zhu}}\email{tingting.zhu@eng.ox.ac.uk}

\author[1,4]{\fnm{David A.} \sur{Clifton}}\email{david.clifton@eng.ox.ac.uk}

\affil*[1]{\orgdiv{Department of Engineering Science}, \orgname{University of Oxford}, \country{UK}}
\affil*[2]{\orgdiv{School of Computer Science}, \orgname{Fudan University}, \country{China}}

\affil[3]{\orgdiv{Department of Computing}, \orgname{Imperial College London}, \country{UK}}
\affil[4]{\orgdiv{Oxford Suzhou Centre for Advanced Research}, \country{China}}

\abstract{
The prevalence of artificial intelligence (AI) has envisioned an era of healthcare democratisation that promises every stakeholder a new and better way of life. 
However, the advancement of clinical AI research is significantly hurdled by the dearth of data democratisation in healthcare.
To truly democratise data for AI studies, challenges are two-fold: 1. the sensitive information in clinical data should be anonymised appropriately, and 2. AI-oriented clinical knowledge should flow freely across organisations. 
This paper considers a recent deep-learning advent, dataset condensation (DC), as a stone that kills two birds in democratising healthcare data.
The condensed data after DC, which can be viewed as statistical metadata, abstracts original clinical records and irreversibly conceals sensitive information at individual levels; nevertheless, it still preserves adequate knowledge for learning deep neural networks (DNNs). 
More favourably, the compressed volumes and the accelerated model learnings of condensed data portray a more efficient clinical knowledge sharing and flowing system, as necessitated by data democratisation.
We underline DC's prospects for democratising clinical data, specifically electrical healthcare records (EHRs), for AI research through experimental results and analysis across three healthcare datasets of varying data types.
}


\maketitle


\section{Introduction}
\label{sec:intro}
The prevalence of artificial intelligence (AI) in healthcare has evolved this field unprecedentedly. 
With its capability to automate physician processes, 
intelligent computing promises to reduce the workloads of clinical practitioners, assist in disease diagnoses and interpretations, improve medical care, foster patient attendance and personalisations, etc. \cite{topol2019high, zhang2022shifting}. 
Pursuing this favourable landscape of providing more predictive, accessible and personalised care drives us to accelerate the deployment of AI systems on clinical grounds. 
For example, the worldwide healthcare AI market is estimated to grow from 19.45 billion USD in 2022 to 280.77 billion USD in 2032, with an annual increase of 30.6\% \cite{ReportsAndData2021a}.
The explosion of intelligent clinical technology has foreshadowed an incoming era of democratised healthcare \cite{Medicine2018}, representing a new and much better way of life for every stakeholder in the health ecosystem.  
Armed with AI, physicians can automate those time-consuming routine tasks and focus on fields that can maximise their values; 
patients can acquire better and personalised care at lower costs; 
analysts can extract and deliver insights from clinical data at unseen scales, just to name a few possible benefits.

Despite the revealed ideal realm of healthcare democratisation, clinical AI research is still challenging and demanding, and the \say{Achilles heel} can be ascribed to a simple but persistent obstacle: the lack of \textit{healthcare data democratisation} \cite{wang2022reinforcing,lewis2020data}. 
Data fuels intelligent algorithms. 
However, healthcare lags behind other industries in data democratisation, particularly data sharing with true openness and interoperability across organisations. 
The digitalisation of medical records, i.e., electronic health records (EHRs) \cite{rajkomar2018scalable}, has engaged patients with an effective way of sharing and managing their own information; however, most EHRs remain isolated from those AI researchers who are keen on devising intelligent systems but without permissions. 
As a result, data democratisation in healthcare is still a Neverland, significantly detrimenting the advance of clinical AI and hurdling healthcare to be truly democratised.

In this work, we are dedicated to placing one crucial puzzle on democratising healthcare data for feeding AI study, and the term \say{data democratisation} refers explicitly to the openness and free sharing of medical record data, particularly EHRs, among clinical AI researchers for developing and deploying intelligent algorithms. 
Challenges, however, are twofold:

\begin{itemize}
\item \textbf{Clincial data is sensitive}. 
The sensitivity of clinical records is a notable hindrance to data democratisation, as they are protected by data regulation laws and cannot be made public without proper anonymisation or de-identifications. 
Anonymisation means no patient can be connected to a specific EHR item. 
A properly anonymised healthcare dataset is not deemed sensitive and can be shared freely without breaking laws or regulations.
In practice, however, the commonly-used healthcare data anonymisation techniques \cite{el2013anonymizing} may not always lead to a truly de-identified dataset, exhibiting an inevitable chance of re-identifying patients across various scenarios \cite{narayanan2010myths,sweeney2015anonymizing,de2015unique}. 
As such, there is an urgent need for an information-sharing system that can deprive the sensitivity of healthcare data but still preserve the knowledge and assets essential to fostering AI innovations. 

\item \textbf{Maintaining, transporting and learning from clinical data can be pricey}. 
For democratic AI study, information flow should be free and open across organisations, necessitating a swift and convenient data-sharing strategy. 
However, the number of EHRs has grown exponentially since healthcare digitalisation, and it can be costly to maintain or transport a large-scale clinical dataset, both economically and time-consuming.
The situation can be further deteriorated by the potential risk of data breaches caused by poor management or hacking. 
Another side effect is that the computational costs for developing intelligent algorithms from such large-scale data can be enormous and thus require the presence of powerful but expensive graphics processing units (GPUs) \cite{pandey2022transformational}, which slows down the trail-and-error cycle of AI research and places hardware barriers against democratisation. 
The high costs of maintaining, transporting, and learning from clinical data necessitate a more efficient and convenient information flow system for medical AI studies.  
\end{itemize}

In this work, we consider recent AI fruition, dataset condensation (DC) \cite{zhao2021distributionmatching}, as a promising candidate for democratising healthcare data. 
DC is a deep-learning technique aiming to learn a \say{condensed} dataset that can preserve the knowledge, in the context of deep model learning, of the original dataset as much as viable. 
This condensed dataset can orient deep neural networks (DNNs) to operate normally on realistic data points without observing any original data of similar patterns. 
Inspiring as it may sound, DC is still an unvisited technique in healthcare, especially regarding its potential for data democratisation.
We spot the two distinctive assets that make it a competent candidate for democratising AI research, i.e.,  the removal of sensitive individual-level information with well-conserved AI utility after DC, and the capability to compress large-scale data for good.

\begin{itemize}
\item \textbf{Condensed data is insensitive metadata and can be freely shared}.
With its unique learning paradigm, DC irreversibly conceals the individual-level information in the original dataset to attain a bona fide de-identification.
In particular, DC aims to match the embedding distributions between the original and condensed data computed by DNNs. 
The parameters of DNNs are not explicitly learnt but are randomly initialised at each training iteration. 
Two random batches of condensed and original data are then selected to learn the former to match its distributions with the latter in the embedding space of those DNNs. 
This paradigm is a stone that kills two birds. 
The knowledge irreversibly flows from original to condensed data via DNNs of randomised parameters operating as innate encrypters.
Moreover, matching random original and condensed batches creates a many-to-one correspondence, i.e., each condensed sample can retain knowledge compressed from numerous original instances.
With the complete removal of individual-level information, linking a patient to a specific EHR in the condensed data is theoretically and computationally infeasible. 
We can view data after DC as statistical data with population-level knowledge sufficient to drive the learning of deep models. 
Since the condensed data is insensitive meta-data, we can freely share it for democratising healthcare AI research without legal concerns. 

\item \textbf{Condensed data can provide a more efficient information flow for AI studies.}
Another notable advantage of DC, as indicated by its name, is that it can significantly compress the size of original EHRs to provide a much faster data transfer between organisations to facilitate interoperability. 
The considerably reduced space occupations also indicate a lower economic cost for maintaining the condensed data. 
More favourably, as the condensed healthcare data does not contain sensitive personal-level information, it can be exempt from the risks of data breaches caused by the re-identifications of patients.
An extra benefit from compressed data volume is that the learning efficiency of DNNs can be significantly accelerated, conveying a much faster convergence rate than the original.
The metrics of DC envision a republic of free and open medical knowledge flow among AI researchers.
\end{itemize}

This work investigates the prospect of applying DC in democratising healthcare AI research for the first time. 
We apply DC to three healthcare datasets of varying clinical variables and tasks to yield condensed versions of train sets. 
We employ a cohort of 11 DNNs of various architectures to evaluate the original and condensed data utilities for deep learning.
We also elaborate on the benefits of condensed data regarding efficient data flow and accelerated model convergence. 
We then discuss the condensed data's inherent metadata nature and illustrate its disparate distribution against the original.
Finally, although still in its infancy, we show that dataset condensation possesses the potential to rescue us from the Groundhog Day of de-identification re-identification races, concentrating on democratising knowledge for healthcare AI research.

\section{Results}
\label{sec:results}
\textbf{Data.} 
We apply DC to three healthcare datasets to examine DC's potential in sharing them. 
The first dataset is denoted as PhysioNet-2012 \cite{silva2012predicting} designed to study the mortality prediction of Intensive Care Unit (ICU) patients, and we analysed the \numprint{8000} publicly-available ICU stays in this work. 
Each ICU stay consists of up to 42 variables collected at least once during the 48 hours after ICU admission.
The target is to predict in-hospital mortalities from those ICU stays.
The second dataset, referred to as MIMIC-III \cite{johnson2016mimic,johnson2016mimic-physionet}, contains \numprint{21156} ICU stays of 17 clinical variables collected from the first 48 hours from entering ICU, and the task is predicting in-hospital death for each stay. 
The last one is named Coswara \cite{sharma2020coswara}, built for studying the coronavirus (Covid-19) diagnosis from various respiratory sounds.
We employ all \numprint{1368} breathing samples and extract acoustic features, aiming to diagnose Covid-19 for each instance. 

Across all datasets, we randomly split all original samples into three sets for training, validation, and tests. 
In PhysioNet-2012, we divide all \numprint{8000} ICU stays into such three sets of \numprint{5120}/\numprint{1280}/\numprint{1600} samples, while we group the \numprint{21156} stays from MIMIC-III into \numprint{14698}/\numprint{3222}/\numprint{3236} data points. Coswara's data split yields three sets of 987/175/206 instances, respectively. 
We thoroughly describe the pre-processing and data splits on three datasets in Supplementary Note 1. \\

\textbf{Condensed Data.}
For each dataset, we learn condensed versions of the train set using 
dataset condensation as described in Section \ref{sec:methods} \say{Methods},  leaving the validation and test sets untouched for future evaluations. 
We generate three condensed sets of varying scales per dataset to illustrate the impacts of the number of condensed samples. 
We condense the train set of PhysioNet-2012 with \numprint{5120} ICU stays into three collections of 20, 40 and 80 data points, respectively. We compress the train set of MIMIC-III (\numprint{14698} ICU
stays) into sets of 400, 800, and \numprint{1200} instances, while we consolidate the 987 training acoustic features of Coswara into groups of 20, 40, and 80 samples. 
We observe DNNs' performances on the test sets when trained on original or condensed sets.  \\

\textbf{The cohort of DNNs.}  
\label{sec:dnn_cohort}
To examine the generality of condensed data to different neural network architectures,  we collect 11 prevalent DNNs for analysing time-series data, most of which have exhibited encouraging performance in the computer vision domain. 
The cohort consists of three multi-scale temporal convolution networks (MS-TCNs) \cite{martinez2020lipreading} with different convolutional kernel sizes, denoted as TCN-\textalpha, TCN-\textbeta, and TCN-\textgamma; 
two vision transformers (ViTs) \cite{dosovitskiy2021an} of distinct attention heads, abbreviated as ViT-\textalpha~and ViT-\textbeta; 
two transformer encoders \cite{vaswani2017attention} of different attention head numbers and dimensions, referred to as TRSF-\textalpha~and TRSF-\textbeta; 
two long short term memory networks (LSTMs) \cite{hochreiter1997long} with varying hidden state units, named LSTM-\textalpha~and LSTM-\textbeta;
and two recurrent neural networks (RNNs) \cite{hopfield1982neural} of diverse hidden dimensions, dubbed RNN-\textalpha~and RNN-\textbeta. 
We involve three networks to learn condensed data: TCN-\textalpha, ViT-\textalpha~and LSTM-\textalpha, while we deploy all 11 DNNs to evaluate the deep-learning utilities preserved in condensed sets.  
We outline the architectures and training settings of those DNNs in Supplementary Note 2. \\

\textbf{Outcomes.} 
The primary outcome is to investigate DC's suitability in democratising healthcare data with its unique condensation mechanism. 
We mainly inspect three facets of this vision: 1). the usability of DNNs on realistic data when trained on condensed data,  2). the benefits introduced by DC in terms of more compact data storage and faster model convergence rates, and 3). the metadata nature of the condensed data.   
The first and the third aspects indicate an unrestricted sharing of condensed healthcare data for AI research, and the second one can encourage an efficient and convenient information flow towards the landscape of data democratisation.
\\

\input{tables/sumup}

\begin{figure}
    \centering
    \includegraphics[width=0.99\textwidth]{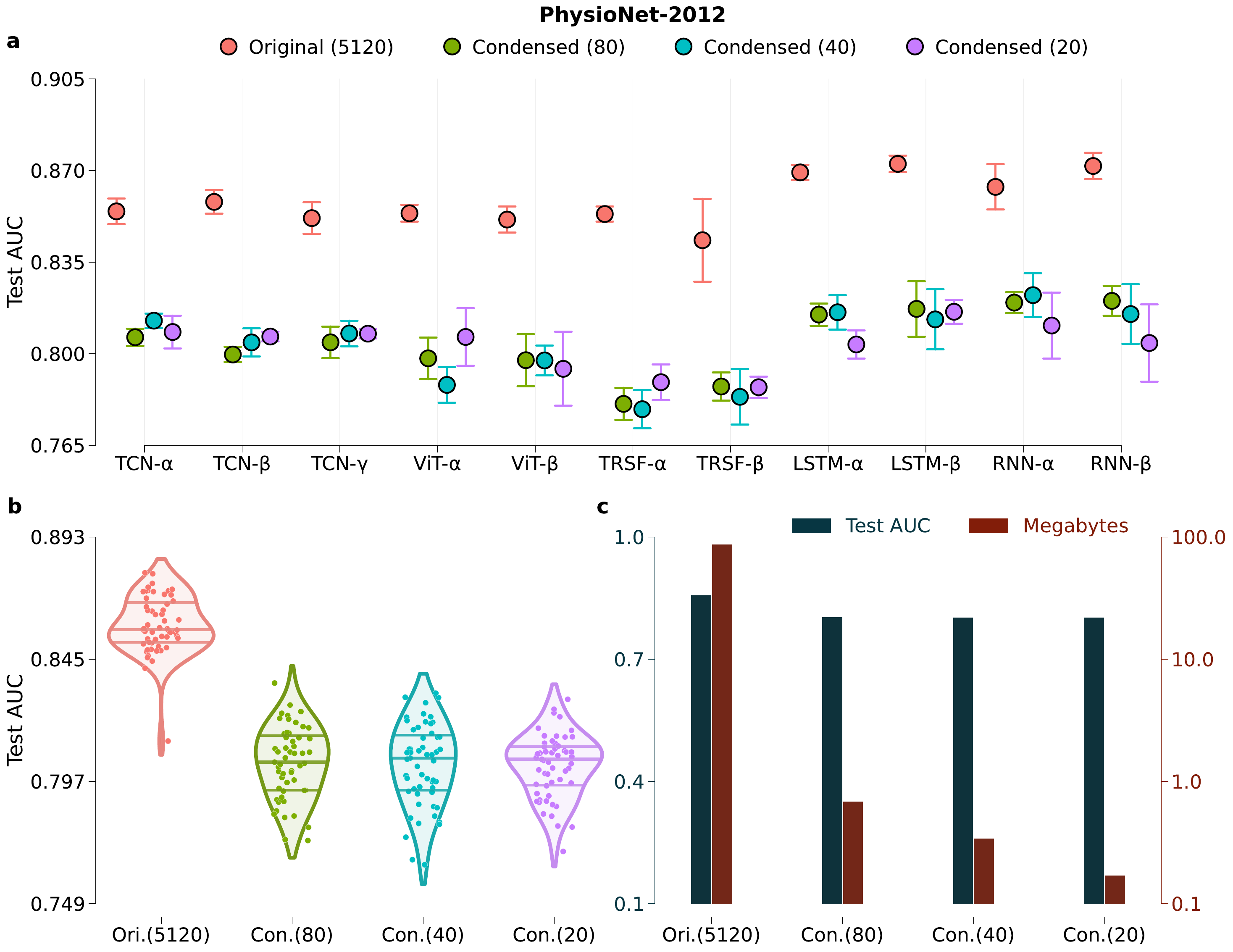}
    \caption{
    \textbf{Visualisations of PhysioNet-2012 results}. \textbf{a}, The average AUCs across 11 DNNs between the original train set and three scales of condensed data. We post-pend the number of samples to each data group. \textbf{b}, The violin plot of all DNNs' test AUCs across the four data groups. \protect\say{Ori.} and \protect\say{Con.} are abbreviations of \protect\say{Original} and \protect\say{Condensed}, respectively. 
    \textbf{c}, Comparisons of the average AUC and disk space occupation across different data groups. Note that we turn on the logarithm scale to display space usage.  
}
        \label{fig: physio_res}
\end{figure}
\begin{figure}
    \centering
    \includegraphics[width=0.99\textwidth]{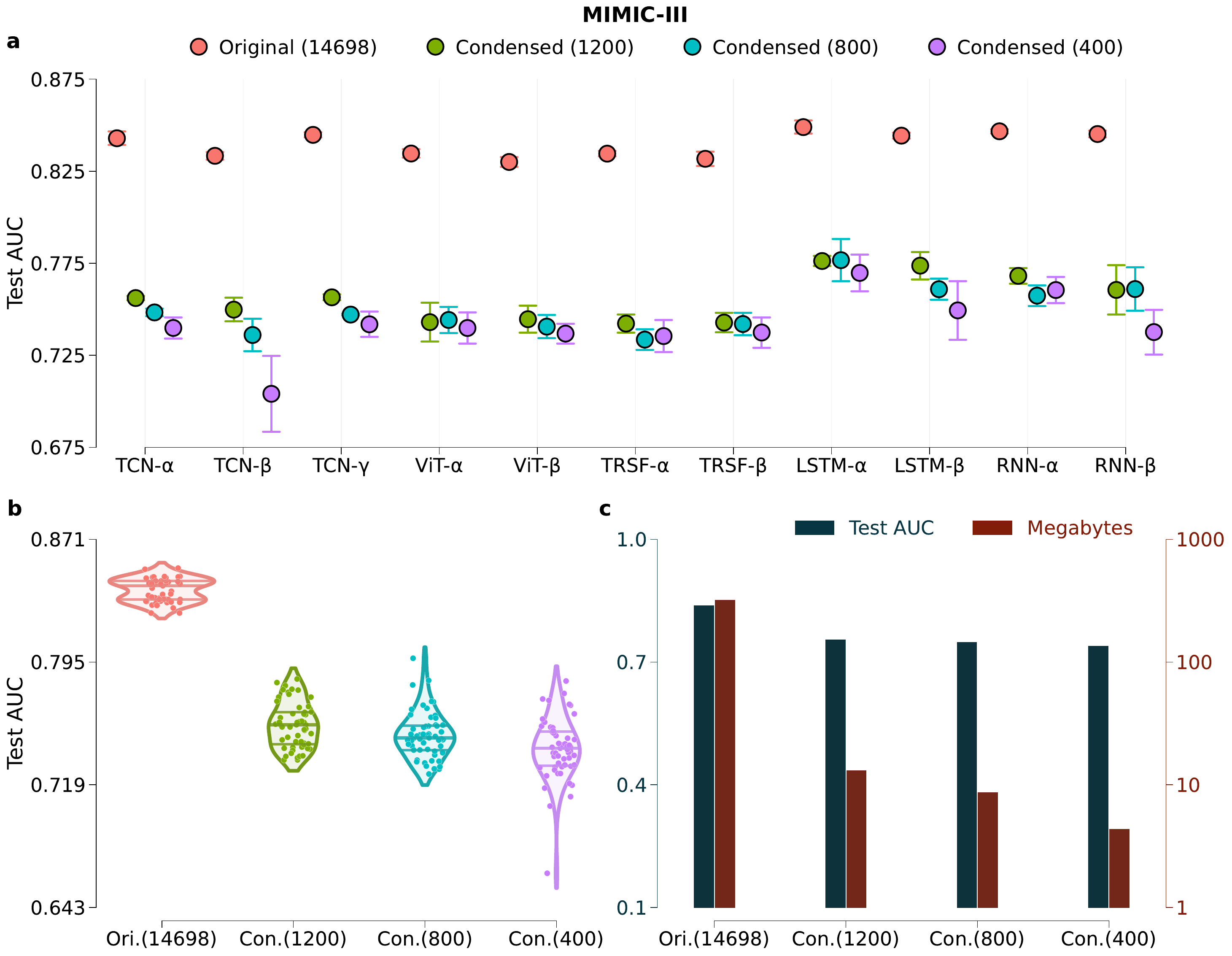}
    \caption{    \textbf{Visualisations of MIMIC-III results}. \textbf{a}, The average AUCs across 11 DNNs between the original and three condensed sets.  \textbf{b}, The violin plot of all DNNs' test AUCs across the four groups.  \textbf{c}, Summarizations of AUCs and data sizes across data groups. We turn on the logarithm scale to display space usage.  }
        \label{fig: mimic3_res}
\end{figure}
\begin{figure}
    \centering
    \includegraphics[width=0.99\textwidth]{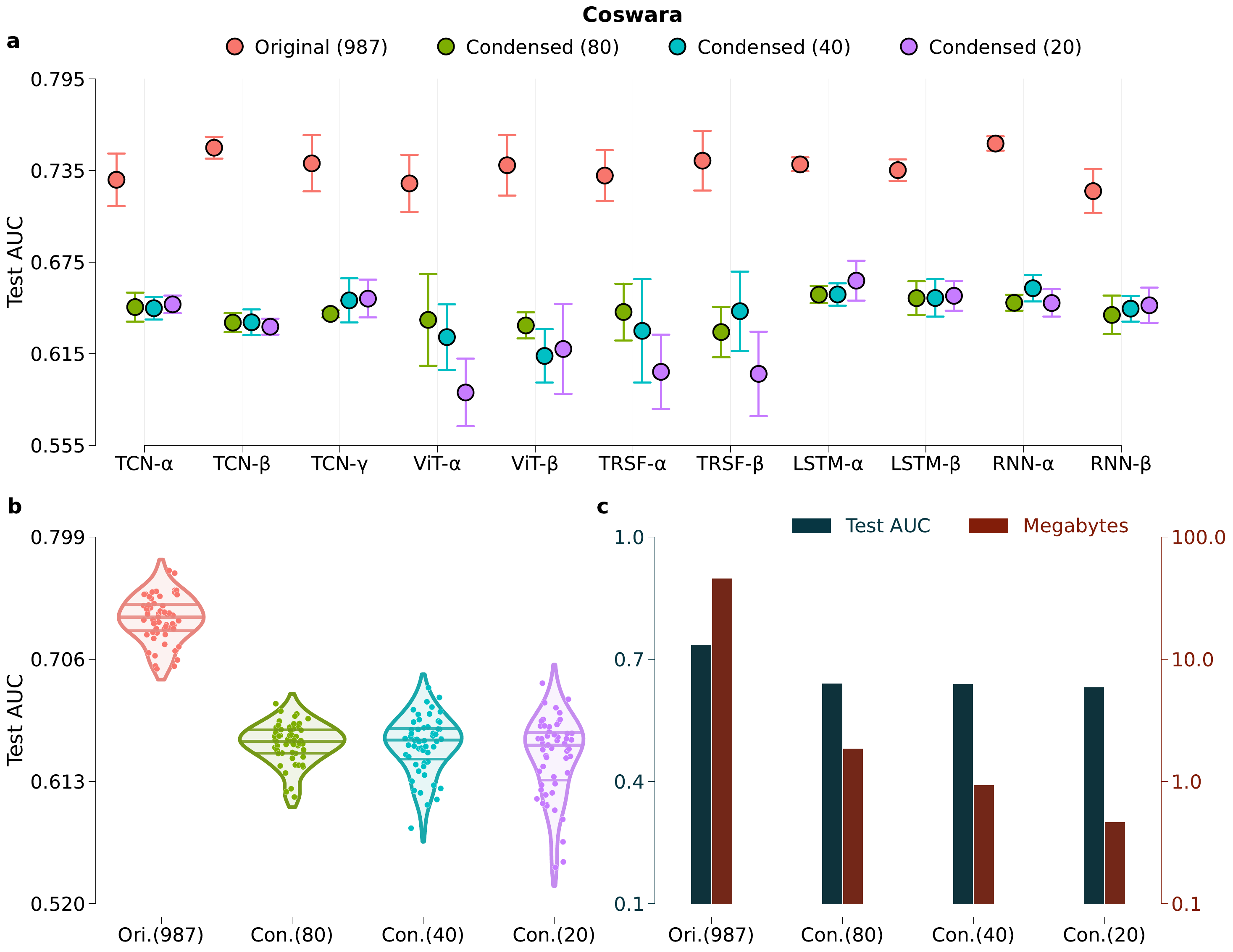}
    \caption{ \textbf{Visualisations of Coswara results}. \textbf{a}, The average AUCs across 11 DNNs between the original and three condensed sets.  \textbf{b}, The violin plot of all DNNs' test AUCs across the four groups.  \textbf{c}, Summarizations of AUCs and data sizes across data groups. We turn on the logarithm scale to display space usage.  }
        \label{fig: covid_b_res}
\end{figure}

\textbf{Condensed data well preserves deep learning utility with significantly compressed volumes.} 
We expect the condensed data to adequately preserve the deep learning utility of original data for DNNs to operate on realistic samples. 
To this end, we employ the widely-used area under the curve (AUC) as the primary evaluation metric. 
We consider the average test AUC of all 11 DNNs trained on the original train sets as the baseline. 
Then we train the same cohort of DNNs on condensed versions of the same set and study how their test AUC alters from the baseline. 
To mitigate bias, we repeat the training of each DNN five times and report the average AUC and standard deviation (SD). 

We summarise the performance of the original and condensed data across three datasets in Table \ref{tab:sum_up}, and we also visualise the results on each dataset in Fig. \ref{fig: physio_res}, Fig. \ref{fig: mimic3_res}, and Fig. \ref{fig: covid_b_res}. 
We enclose the numeric results per DNN in Supplementary Table 8. 
As shown in Table \ref{tab:sum_up} and Fig. \ref{fig: physio_res}c, the original train set of PhysioNet-2012 with \numprint{5120} ICU stays can teach all 11 DNNs to attain an average test AUC = 0.858 (SD 0.011) of predicting in-hospital mortality. Nevertheless, it occupies 88.13 megabytes (MBs) of disk space. 
On the other hand, DNNs trained on the condensed data of 80, 40, and 20 samples exhibit slightly reduced AUCs = 0.858 \textrightarrow~0.804, 0.804, and 0.803, respectively, reflecting an approximate 5.5\% drop. 
Such accuracy costs are arguably acceptable, or even impressive, given the significantly compressed data sizes = 88.13 \textrightarrow~0.69, 0.34, and 0.17 MBs, deducting 87.44 MBs (99.2\% less than the original) even at the worst case of 80 samples. 
Besides, the AUCs across three condensed groups are parallel and stable, disregarding the varying number of samples, although the sets of 80 and 40 samples slightly outperform the rest with 20 points. 
We can also read an interesting phenomenon from the performance comparison of individual DNNs in Fig. \ref{fig: physio_res}a.
A higher AUC of a certain DNN learnt on the original data will usually accompany better AUCs of the same architecture on the condensed data and vice versa. 
This tendency can be in line with the mechanism of DC, i.e., matching the embedding distributions between the original and condensed data batches, which will likely result in similar embedding patterns for the same DNN structure. 
In addition, although DNNs other than TCN-\textalpha, ViT-\textalpha~and LSTM-\textalpha~remain unused during the learning of DC, they can still gain comparable or even better AUCs on the condensed data than the visited three, implying DC's generalisation capability to a wide variety of neural network architectures. 

We can also discover evidence of preserved deep-learning utility from condensed data from MIMIC-III results. 
As displayed in Table \ref{tab:sum_up} and Fig. \ref{fig: mimic3_res}c, the 11 DNNs learnt on MIMIC-III's original train set of \numprint{14698} ICU stays can attain a mortality prediction of AUC = 0.840 (SD 0.007), whereas 322.95 MBs space are needed for harbouring the data.
Training the same DNNs collection with condensed sets of 1200, 800, and 400 elements yields lower test AUCs = 0.840 \textrightarrow~0.756, 0.750, and 0.741, respectively. 
This 8.9\% drop in prediction accuracy can be effectively compensated by the considerably relieved storage requirement = 322.95 \textrightarrow~13.18, 8.79, and 4.39 MBs, which occupies 95.9\%, 97.2\%, and 98.6\% less space, respectively. 
The AUC results of individual DNNs in Fig. \ref{fig: mimic3_res}a demonstrate a stable AUC correlation between the original and condensed data on the same DNN architectures, a phenomenon consistent with the PhysioNet-2012 results. 

On Coswara, we apply DC to a distinct clinical scenario, namely diagnosing Covid-19 from acoustic features of breathing sounds. 
Considering the achieved DNN accuracy, this task is more formidable than the mortality prediction on PhysioNet-2012 or MIMIC-III. 
All 11 DNNs learnt with the original train set of 46.27 MBs, as reported in Table \ref{tab:sum_up} and Fig. \ref{fig: covid_b_res}c, can only acquire an average test AUC = 0.737 (SD 0.017), which is less satisfying than that of PhysioNet-2012 or MIMIC-III. 
Constrained by the AUCs from the original data, the condensed groups of 80, 40, and 20 instances orient DNNs to achieve AUCs = 0.737 \textrightarrow~0.642, 0.641, and 0.632, a drop of around 9.6\% AUC. 
DC compresses the disk usage from 46.21 MBs of original into 1.88, 0.94, and 0.47 MBs (80, 40, and 20 condensed instances), respectively, saving 96.0\% space. 
Fig. \ref{fig: covid_b_res}a depicts the AUC comparisons of each architecture, and the overall trend is consistent as in PhysioNet-2012 or MIMIC-III.

Across three datasets, we can observe that the deep-learning utility of the original data, represented by the average AUC of 11 DNNs, is vastly kept in the condensed data across varying clinical scenarios, despite the consolidated volumes that facilitate the data storage and transfer.
Notably, the generality of dataset condensation to unseen neural architectures also raises the values of DC in data democratisation.   
\\

\begin{figure}
    \centering
    \includegraphics[width=0.99\textwidth]{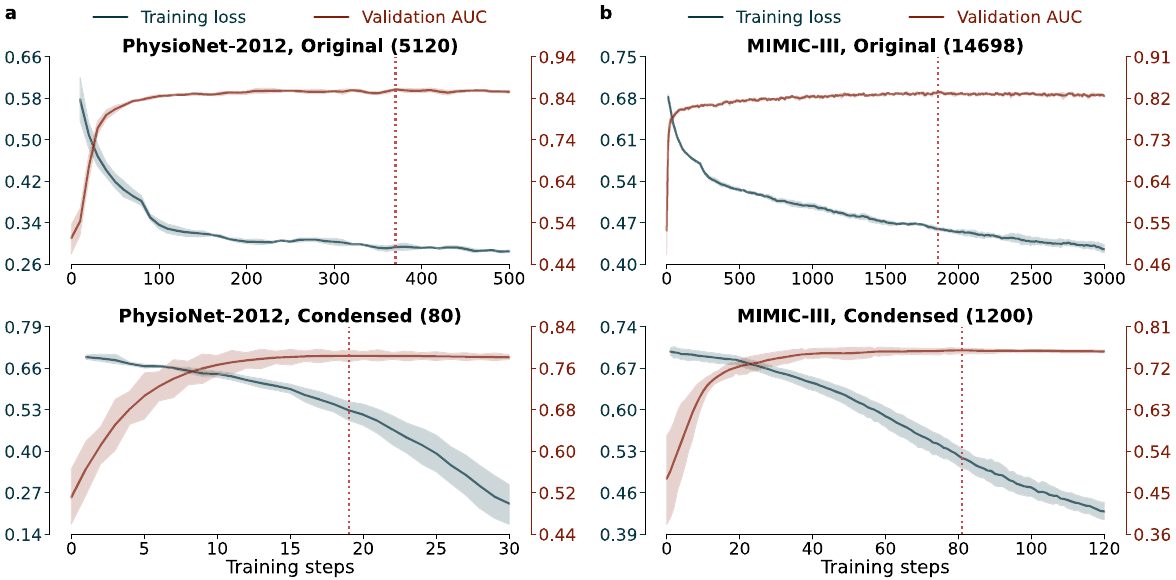}
    \caption{\textbf{a}, The learning curves of TCN-\textbeta~trained on the original and condensed data of PhysioNet-2012. \textbf{b}, The learning curves of TRSF-\textalpha~trained on the original and condensed data of MIMIC-III. The red-dotted line indicates the step of convergence. } 
    \label{fig: learning_curve}
\end{figure}

\input{tables/training_steps.tex}

\textbf{Condensed data can lead to a revved convergence of deep models.} 
Sharing the condensed data can introduce another attractive property to AI research: it can accelerate the convergence of DNNs and thus alleviate time costs. 
To describe this benefit, we plot the learning curves of TCN-\textbeta~trained on PhysioNet-2012's original and condensed set in Fig. \ref{fig: learning_curve}a, and the learning curves of TRSF-\textalpha~on MIMIC-III in Fig. \ref{fig: learning_curve}b, respectively. 
We can observe from Fig. \ref{fig: learning_curve}a that on PhysioNet-2012, it takes approximately 370 training steps for TCN-\textbeta~ to attain convergence (denoted by the red dotted line) on the original train set. 
The same network trained on 80 condensed samples converges at step = 370 \textrightarrow~19, saving 93.1\% of training time.    
Likewise, as shown in Fig. \ref{fig: learning_curve}b, the average time for TRSF-\textalpha~to converge in the original MIMIC-III train set is \numprint{1860} training steps, while substantially fewer convergence steps = \numprint{1860} \textrightarrow~81 are required on the condensed set of 1200 instances,  which accelerates the training session by 95.6\%. 
We enclose the learning curves of addition networks in Supplementary Note 3.

We deliver a detailed comparison of the convergence steps between the original and condensed data of each DNN in Table \ref{tab:training_steps}.
It can be discovered that accelerated learning of condensed data is a common trait across all datasets and networks. 
On average, the condensed data can boost the training speed by 95.4\%, 91.1\%, and 88.0\% than the original on PhysioNet-2012, MIMIC-III, and Coswara, respectively.
Consequently, sharing condensed data can benefit healthcare AI research with significantly compacted volume and faster model learning. Such traits can be highly desirable when studying a dataset of massive data volumes that can be time-consuming to complete the model training.
\\

\textbf{Condensed data effectively conceals individual-level knowledge.} 
The elimination of individual-level information is probably the most attractive property of DC, aside from the well-preserved ML utility and compressed data volume. 
As described in Section \ref{sec:methods}, DC performs distributions matching between two random batches of original and condensed data embedded by randomised DNNs. 
Since the number of condensed samples is far less than the original, this mechanism creates a many-to-one correspondence. 
Each condensed sample receives knowledge from multiple original instances, and the condensed set is essentially a macro-level synthesis of the original dataset.
From this perspective, sharing a condensed dataset is similar to sharing the metadata, such as the mean and STD of clinical variables, which hides individual-level information and is thus permitted by relevant regulations. 

\begin{figure}
    \centering
    \includegraphics[width=0.99\textwidth]{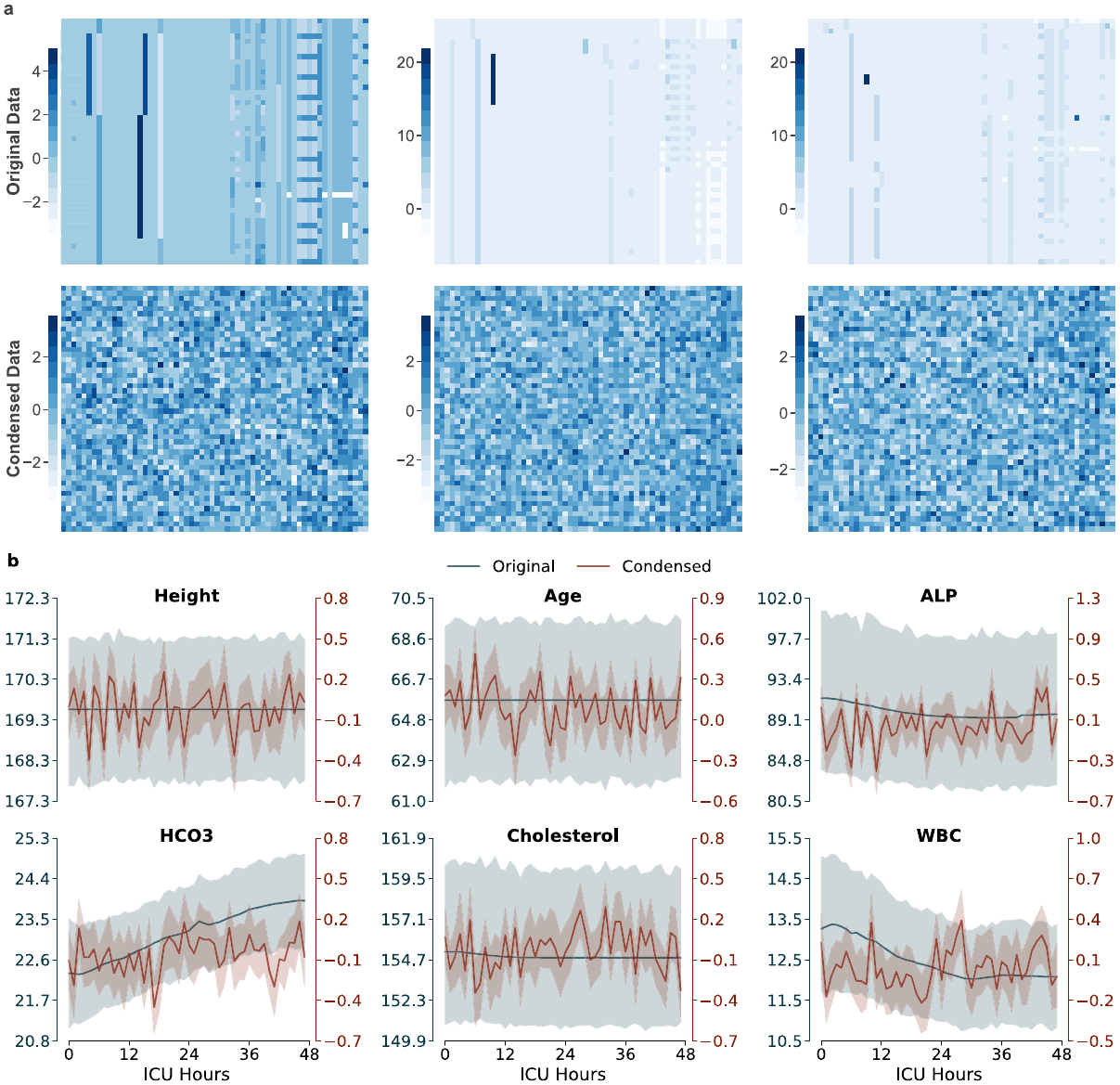}
\caption{Visualisations of original and condensed samples. 
\textbf{a}, the heatmaps of three original and three condensed samples randomly drawn from MIMIC-III. 
\textbf{b}, the 48-hour trends of six clinical variables from PhysioNet-2012 computed from 80 original and 80 condensed instances, respectively. 
}
        \label{fig: data_comparsion}
\end{figure}

To illustrate this metadata property, we visualise in Fig. \ref{fig: data_comparsion}a six data examples randomly chosen from the MIMIC-III and its condensed set as 2D heatmaps. 
The heatmaps of original instances exhibit highly regulated and potentially identifiable patterns, representing a potential threat of patient re-identifications.
In contrast, the heatmaps of condensed samples display patterns of complete chaos, indicating that they are essentially random signals without any explicit meaning. 
Those unorganised signals result from the learning of DC; each condensed instance receives and fuses the knowledge of multiple original samples.
As a result, original data of organised patterns are entangled and randomised into samples of chaotic signals, which can arguably conceal the individual-level information in healthcare data. 

We further investigate this issue by examining various clinical variables. 
Fig. \ref{fig: data_comparsion}b depicts the 48-hour trends of six clinical variables computed from 80 original and 80 condensed samples of PhysioNet-2012.
Thoe variables are: \say{height}, \say{age}, \say{alkaline phosphatase} (\say{ALP}), \say{serum bicarbonate} (\say{HCO3}), \say{cholesterol}, and \say{white blood cell count} (\say{WCB}).
As general descriptors, \say{height} and \say{age} from the original samples demonstrate constant values across the ICU period. 
However, the same variables from the condensed data show significantly fluctuating trends over time, which is improbable in real cases and can shield the variable-level information. 
We can observe similar deviations between the rest original and condensed variables. 
For instance, the overall value of the original \say{ALP} and \say{WCP}  declines as hours increase, while their condensed curves display an inconsistent trend with randomised noises. 
Fluctuations on condensed variables can be discovered on \say{HCO3} with an increasing tendency in the original.
We can notice that irregular fluctuations of variables are common traits of condensed samples, despite the conceivably regular patterns in the original variables. 
Such observations are consistent with the heatmap visualisation of Fig. \ref{fig: data_comparsion}a and reflect an unrecoverable personal knowledge concealment at the clinical variable level.
Besides, the range of each condensed variable is automatically shifted and rescaled into arbitrary fields approximately between $-1$ and $1$, which further encrypts variable-level knowledge.  
We supplement more comparisons of original and condensed samples in Supplementary Note 3.
\\

\textbf{The distribution of condensed data is disparate from the original.}
\begin{figure}
    \centering
    \includegraphics[width=0.99\textwidth]{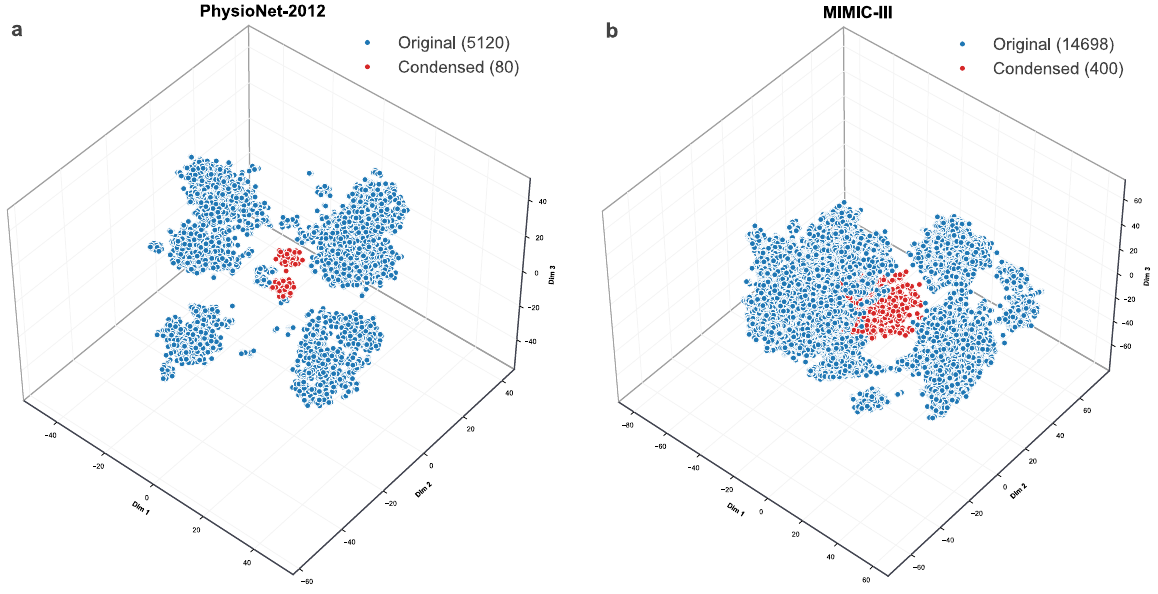}
    \caption{The visualisations of the original and condensed samples embedded into 3D with t-SNE.  \textbf{a}, the 3D distributions of \numprint{5120} original and 80 condensed samples of PhysioNet-2012. \textbf{b}, the 3D distributions of \numprint{14698} original and \numprint{400} condensed samples of MIMIC-III.  }
        \label{fig: tsne_vis}
\end{figure}
Concealments at individual and variable levels imply a potential disparate distribution in condensed data that effectively hides the realistic data distribution.
Sharing condensed data is, therefore, sharing an unseen dataset independent from the original.
To explain this distribution shifting, we apply t-distributed stochastic neighbour embedding (t-SNE) \cite{van2008visualizing} to visualise the 3D distributions of the original and condensed data of PhysioNet-2012 in Fig. \ref{fig: tsne_vis}a and MIMIC-III in Fig. \ref{fig: tsne_vis}b.
PhysioNet-2012's original data points loosely constitute four distant clusters as in Fig \ref{fig: tsne_vis}a, leaving an unoccupied heart region.
On the other hand, the condensed points exhibit a deviating distribution. 
Most reside within the central but empty region separating the original clusters, and we observe only two distinctive groups. 
Such distribution can result from the many-to-one correspondence in DC. Since each condensed point receives information from multiple original samples, their distribution naturally moves to the central region that best balances the knowledge from different cases. 
From this perspective, the condensed data substantively captures the macro statistics of the original dataset, and sharing them is just like sharing statistical data such as mean or STD. 
We can see from Fig \ref{fig: tsne_vis}b that the distributions between the original and condensed on MIMIC-III also align with this analysis.
Despite the significantly increased density, MIMIC-III's original data constitute several clusters with a sparse central area in which all condensed points reside, indicating the metadata essence of condensed data.

\begin{figure}
    \centering
    \includegraphics[width=0.9\textwidth]{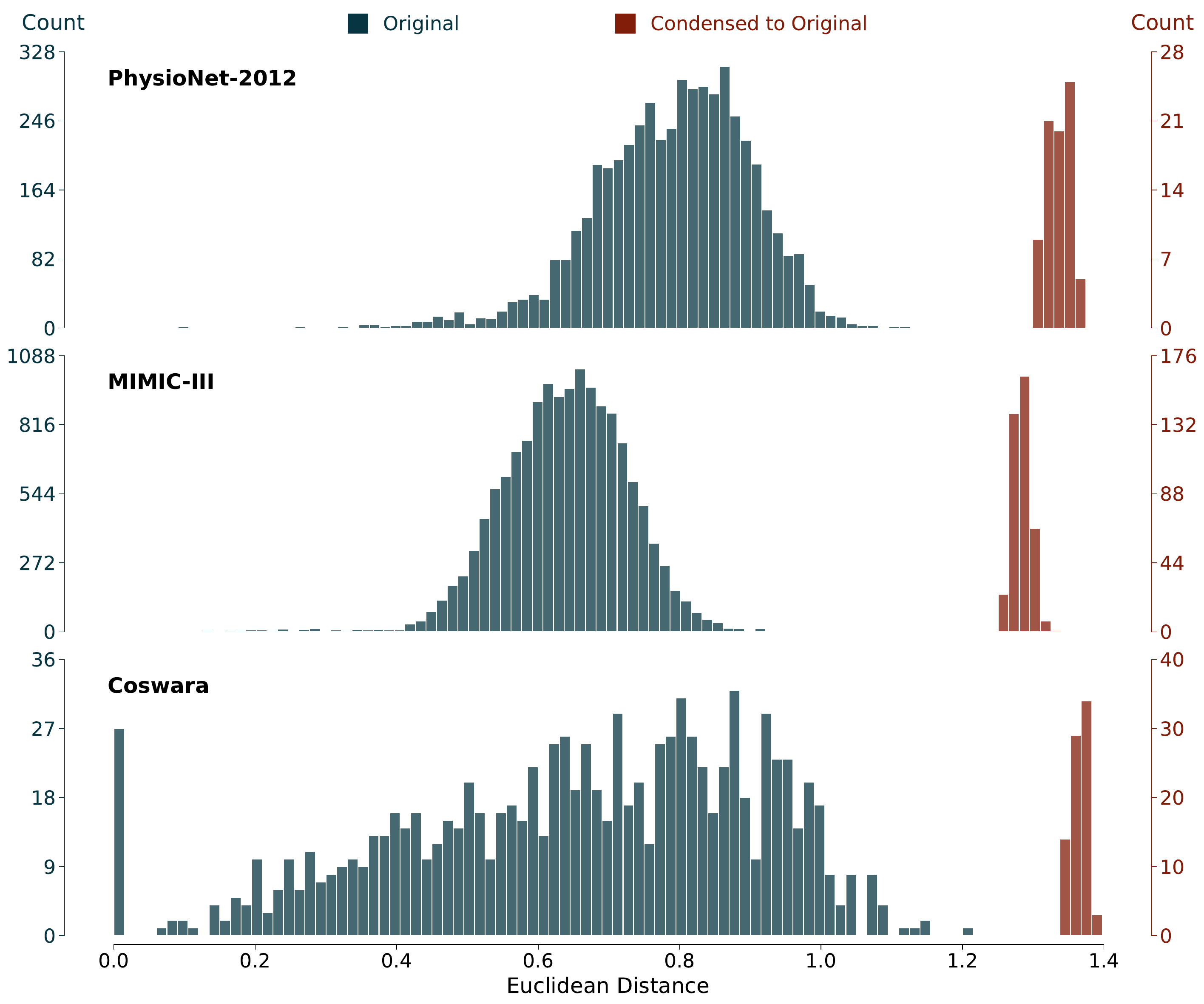}
    \caption{The distance histograms of PhysioNet-2012, MIMIC-III, and Coswara. 
    We find each condensed sample's minimum Euclidean distance with every original instance and plot the minimum distances of all condensed samples as the histogram (\protect\say{Condensed to Original} in red).
    For each original sample, we find its minimum Euclidean distance with every other instance and plot the histogram for all original ones (\protect\say{Original} in blue).
    }
        \label{fig: dist_hist}
\end{figure}

We further inspect the distribution shifting by computing Euclidean distances between samples. 
We plot in Fig. \ref{fig: dist_hist} the histogram of Euclidean distances within the original train sets and the distance histogram from the condensed instances to the original, respectively.
Across all three datasets, the distances within the original data are significantly less than those from condensed to the original. 
For example, the distances between most MIMIC-III original samples fall between $0.4$ and $0.9$, while the distances from condensed instances to the original ones significantly increased to the scope of $1.3$ and $1.4$. 
In other words, any condensed-to-original distance is around two times longer than the original-to-original, indicating that condensed points inhabit space distant from the original span. 
Similar distribution divergences can also be observed from the histograms of PhysioNet-2012 and Coswara in Fig. \ref{fig: dist_hist}, in which the distances between condensed and original samples significantly surpass those between original ones. 
With the shifted distribution, democratising condensed data can be free of possible patient re-identifications that persist in sharing individual-level data of realistic distributions. 

\section{Discussion}
\label{sec:discussion}
AI research stands at the heart of healthcare democratisation; however, the lack of a satisfactory data democratisation mechanism is a long-haunted issue that detriments clinical AI development.
This paper examines the potential of dataset condensation, a recent AI fruition, in democratising medical data for AI research.
The experimental results on three healthcare datasets, PhysioNet-2012,  MIMIC-III, and Coswara, reveal that DC can preserve the deep-learning utility in the original clinical records and simultaneously facilitate information flow with the compressed data volume. 
Condensed data can also teach DNNs to converge at much faster paces, significantly accelerating the learning of deep networks.
The metadata nature of condensed data is also analysed and discussed, which irreversibly conceals individual-level information with the many-to-one learning mechanism and disparate distributions. 
As population-level knowledge, like statistical metadata, condensed data can be shared and democratised to fuel medical AI research. 
The prospect of DC is promising. 
It may free us from the unceasing races between de-identifications and re-identifications, focusing on democratising AI-oriented knowledge for deep learning research.
Data holders of clinical datasets of highly sensitive but also valuable knowledge can apply DC to create a condensed set to be freely shared without the risk of individual re-identifications. 
Healthcare researchers can also share the condensed version of their private datasets that are usually time-consuming or impossible to receive access approvals. 
Other researchers can use condensed datasets of significantly compressed sizes to train less accurate but still workable DNNs at accelerated convergence rates. 
Although still in its infancy, DC can present inspiring outlooks of facilitating scientific studies' openness and democratising AI research in healthcare. 

On the other hand, dataset condensation has various limitations, and we elaborate on them with a \textit{no free lunch} view. 
Any de-identification method for data sharing comes at a cost, typically impaired data utility. 
Classic anonymisation approaches attain de-identification by redacting direct and quasi-identifiers, leading to a degradation in the data utility. 
Removing more data fields will generally lead to a lower chance of re-identification but at the cost of further demolished usabilities. 
This trade-off between de-identification and data utility can also be applied to sharing synthetic data acquired with generative adversarial networks (GANs) \cite{goodfellow2020generative,xu2019modeling}. 
Better sensitive information protections in synthetic data come at the price of deteriorated usability, and vice versa for retaining good data utility and increased vulnerability to patient re-identifications. 
Besides, drawn from a realistic data distribution, synthetic data possesses individual-level information of original samples that may reveal personal identifications.

The \textit{no free lunch} view also retains for DC. Condensed data is essentially macro-level statistical descriptors of the original dataset, effectively concealing sensitive information at the individual level. 
However, such concealments also eliminate data usabilities at fine-granted scales crucial to exploiting the full value of clinical records. 
Condensed data can be of limited profit for analytical tasks that demand access to microdata, and this is one of the major costs for free sharing. 
Also, the compressed data volume that facilitates storing and model learning docks at the expense of inevitable information losses, represented by the degraded prediction accuracy in learned DNNs. 
There is no free lunch in data democratisation; we must review the pros and cons simultaneously.
Despite the annihilated individual-level utility, the most attractive merit of DC is that it permits the free sharing and flow of AI-oriented knowledge, which could potentially lead to a democratic healthcare system.
Healthcare AI research can profoundly benefit from democratising condensed data of those clinical datasets that are usually difficult to access directly.  

We consider dataset condensation as a newly emerged and encouraging data-sharing route for deploying healthcare AI.
There were approximately three categories of de-identification mechanisms before DC: 1). the classic redaction-based data anonymisation \cite{el2013anonymizing}, 2). the GAN-generated synthetic data sharing \cite{stadler2022synthetic}, and 3). remote-processing systems for delivering population-level insights such as secure multiparty computation or homomorphic encryption \cite{froelicher2021truly}.
DC reveals an unvisited avenue with unique advantages. 
Individual-level knowledge within anonymised or synthetic data represents an inevitable chance of re-identifying patients. 
They also face the quagmire of finding a satisfying trade-off between security and utility.
On the other hand, DC eliminate individual-level knowledge and represents greater sharing flexibility, and applying DC is also more straightforward without security-utility balancing steps. 
DC also differs from those remote-processing systems in multiple aspects. 
Building those remote systems can be costly and time-consuming, including hardware setup, daily maintenance, framework construction, etc.
After being built, those systems could only support pre-defined analytical tasks and lacked the flexibility to adapt to learning models of varying architectures. 
DC, in contrast, is a low-cost and easy-to-implement technique, and we can deploy it to output condensed data directly and swiftly (see Supplementary Table 10 for the required learning time).
The well-preserved deep learning utility in condensed data also significantly distinguishes DC from remote systems like homomorphic encryption.

To sum up, no silver bullet can simultaneously attain perfect de-identification and flawless data utility. 
Dataset condensation, however, can be a promising technique for sharing and distributing AI-oriented knowledge to benefit healthcare AI research with its unique properties. 
Future works can be devoted to further evolving DNNs' performance learned with condensed data while maintaining other desired traits;
true healthcare democratisation may not be as elusive as it used to seem with DC in vision.

\section{Methods}
\label{sec:methods}
\textbf{Problem definition.}
We define the original train set of a healthcare dataset as $\mathcal{O}=\{(\mathbf{x}_i, y_i) \mid 1 \leq i \leq \abs{\mathcal{O}}\}$ where $\mathbf{x}_i \in \mathbb{R}^{T \times F}$ is the $i$-th time-series sample with $T$ and $F$ as the temporal and feature dimension, respectively, $y_i$ is the class label, and $\abs{\mathcal{O}}$ refers to the total number of samples.
We aim to learn a condensed dataset $\mathcal{C}=\{(\mathbf{c}_i, y_i) \mid 1 \leq i \leq \abs{\mathcal{C}}\}$
where $\mathbf{c}_i \in \mathbb{R}^{T \times F}$ is a condensed sample, $\abs{\mathcal{C}}$ is the number of condensed samples, and $\abs{\mathcal{C}} \ll \abs{\mathcal{O}}$.
We expect that $\mathcal{C}$ can preserve the deep-learning utility in $\mathcal{O}$ such that when replacing the train set $\mathcal{O}$ with $\mathcal{C}$, DNNs can still achieve parallel performance on an unseen test set, which can be denoted as 
\begin{linenomath}
\begin{equation}
\label{eq: problem}
\mathbb{E}_{\mathbf{x} \sim P_{\mathcal{D}}} [\fnc{L}(\varphi_{\bm{\theta}^{\mathcal{O}}}(\mathbf{x}), y)] \simeq 
\mathbb{E}_{\mathbf{x} \sim P_{\mathcal{D}}} [\fnc{L}(\varphi_{\bm{\theta}^{\mathcal{C}}}(\mathbf{x}), y)]. 
\end{equation}
\end{linenomath}
In Eq. \ref{eq: problem}, $\mathbf{x} \in \mathbb{R}^{T \times F}$ is a data sample drawn from the original data distribution $P_{\mathcal{D}}$, $y$ is the class label, $\fnc{L}$ is a loss function such as the binary cross-entropy (BCE), $\varphi$ is a DNN with parameters $\bm{\theta}$,  
$\varphi_{\bm{\theta}^{\mathcal{O}}}$ and $\varphi_{\bm{\theta}^{\mathcal{C}}}$ denote the DNN trained on $\mathcal{O}$ and $\mathcal{C}$, respectively.
Eq. \ref{eq: problem} is DC's objective for time-series healthcare data; however, it is challenging to optimise this equation directly. 
We follow \cite{zhao2021distributionmatching} to solve Eq. \ref{eq: problem}, seeking to match the distributions in embedding spaces which are more efficient to compute. \\
 
\textbf{Dataset condensation with distribution matching.}
The major challenge of optimising Eq. \ref{eq: problem} is the estimation of data distribution $P_{\mathcal{D}}$. Considering the high-dimensional healthcare samples, obtaining an accurate and reasonable $P_{\mathcal{D}}$ can be computationally expensive and practically infeasible. 
To address the same issue in computer vision, authors of \cite{zhao2021distributionmatching} assume that each image sample can be embedded into a lower-dimensional space with a group of parametric functions and then assemble results to estimate the original image data distributions.
In this work, we obey the same premise for healthcare data, embedding each sample $\mathbf{x} \in \mathbb{R}^{T \times F}$ with a collection of parametric functions $\phi_{\bm{\mu}}: \mathbb{R}^{T \times F} \mapsto \mathbb{R}^{T^{\prime} \times F^{\prime}}$ where $T^{\prime} \ll T$, $F^{\prime} \ll F$, and $\bm{\mu}$ stands for the parameters of a function $\phi$. 
Each $\phi_{\bm{\mu}}$ reveals a fragment of the original sample distribution while merging them together can provide a reasonable $P_{\mathcal{D}}$ estimation.

Under this landscape, we can match embeddings between the original and condensed datasets using the maximum mean discrepancy (MMD) distances \cite{gretton2012kernel}, which can be described as 
\begin{linenomath}
\begin{equation}
\label{eq: MMD_distance}
\sup_{\norm{\phi_{\bm{\mu}}}_{\mathcal{H}} \leq 1} (\mathbb{E}[\phi_{\bm{\mu}}(\mathcal{O})] - \mathbb{E}[\phi_{\bm{\mu}}(\mathcal{C})])
\end{equation}
\end{linenomath}
where $\mathcal{H}$ represents the reproducing kernel Hilbert space (RKHS). 
We employ an empirical estimation of MMD distance in Eq. \ref{eq: MMD_distance} following \cite{zhao2021distributionmatching}, which can be written as 
\begin{linenomath}
\begin{equation}
\label{eq: MMD_estimate}
\mathbb{E}_{\bm{\mu} \sim P_{\bm{\mu}}} \norm*{
\frac{1}{\abs{\mathcal{O}}} \sum_{i=1}^{\abs{\mathcal{O}}} \phi_{\bm{\mu}}(\mathbf{x}_i) - \frac{1}{\abs{\mathcal{C}}} \sum_{j=1}^{\abs{\mathcal{C}}} \phi_{\bm{\mu}}(\mathbf{c}_j)
}{^{\mathclap{2}}}
\end{equation}
\end{linenomath}
where $P_{\bm{\mu}}$ refers to the parameter distribution of $\phi_{\bm{\mu}}$. 

There are different methods for embedding networks $\phi_{\bm{\mu}}$. The authors of \cite{zhao2021distributionmatching} took a fixed network architecture and sampled randomised parameters at each training step. 
We extend this approach to a group of network architectures in pursuit of better structural generalisations and de-identifications. 
For each training step, we first randomly pick a network  $\phi_{\bm{\mu}_{i}}^{i}$ from a collection $\Phi = \{ \phi_{\bm{\mu}_{i}}^{i} \mid 1 \leq i \leq \abs{\Phi} \}$ where $\phi_{\bm{\mu}_i}^{i}$ denotes the $i$-th DNN with parameters $\bm{\mu}_i$ and $\abs{\Phi}$ denotes the collection size. We generate randomised weights from parameter distribution $P_{\bm{\mu}_i}$ for the selected network.
This dual-sampling strategy turns the objective into an optimisation issue that can be described as
\begin{linenomath}
\begin{equation}
\label{eq: objective}
\min_{\mathcal{C}} \mathbb{E}_{\phi_{\bm{\mu}_k}^k \sim \Phi, \bm{\mu}_k \sim P_{\bm{\mu}_k}} \norm*{
\frac{1}{\abs{\mathcal{O}}} \sum_{i=1}^{\abs{\mathcal{O}}} \phi_{\bm{\mu}_k}^k (\mathbf{x}_i) - \frac{1}{\abs{\mathcal{C}}} \sum_{j=1}^{\abs{\mathcal{C}}} \phi_{\bm{\mu}_k}^k (\mathbf{c}_j)
}{^{\mathclap{2}}}.
\end{equation}
\end{linenomath}
Eq. \ref{eq: objective} states that we would like to learn a condensed dataset $\mathcal{C}$ that can minimise the discrepancy between the two distributions of original and condensed data in the embedding spaces projected by an arbitrary network $\phi_{\bm{\mu}_k}^k$ with randomly-sampled parameters $\bm{\mu}_k$.

The randomised network architecture and parameters are probably one of the most interesting mechanisms of DC.
At first glance, it may be counter-intuitive how a randomised network of randomised parameters can lead to a proper embedding of original data, and we may also suspect how the deep-learning knowledge can flow from the original to the condensed dataset smoothly. 
Relevant study \cite{saxe2011random}, however, indicates that randomly-initialised networks can extract feature descriptors of high qualities for various computer vision tasks.  
Those networks can also maintain the distance affinity of original samples in the embedding spaces, e.g., the embedding distance of two samples of the same category is less than those of different classes, as demonstrated in \cite{giryes2016deep}. 
With high-quality and distance-preserving representations, we can expect randomly-initialised DNNs to construct a meaningful distribution of original data in the embedding space, which can adequately guide the learning of the condensed dataset.
Applying this randomised-network strategy can also yield a favourable trait in healthcare: the irreversible knowledge flow from original to condensed data. 
It is computationally impossible to trace back to any original data from the condensed dataset without knowing network $\phi_{\bm{\mu}_k}^k$ and sampled parameters $\bm{\mu}_k$ that are forgotten after training of DC. \\

\input{tables/learning_algorithm.tex}

\textbf{Learning the condensed dataset.} To learn the condensed dataset $\mathcal{C}$ in Eq. \ref{eq: objective}, we employ a mini-batch training algorithm as in \cite{zhao2021distributionmatching}. 
We first randomly initialised a condensed dataset $\mathcal{C}$, which is learnt for a total of $M$ iterations. For each iteration, we perform randomised samplings as follows: 1). sample a network $\phi_{\bm{\mu}_k}^k \sim \Phi$, 2). sample this network's parameters $\bm{\mu}_k \sim P_{\bm{\mu}_k}$, 3). for each class $s$, sample an original data batch $B^{\mathcal{O}}_s \sim \mathcal{O}$ and a condensed data batch $B^{\mathcal{C}}_s \sim \mathcal{C}$. The average discrepancy between original and condensed data batches across all classes is computed as the loss $\mathcal{L}$ at this iteration, which can be written as
\begin{linenomath}
\begin{equation}
\label{eq: final_loss}
\mathcal{L} = \sum_{s=1}^{S} \norm*{
\tfrac{1}{\abs{B^{\mathcal{O}}_s}} 
\sum_{\mathclap{(\mathbf{x}, y) \in B^{\mathcal{O}}_s}} \phi_{\bm{\mu}_k}^k(\mathbf{x})
- \tfrac{1}{\abs{B^{\mathcal{C}}_s}} 
\sum_{\mathclap{(\mathbf{c}, y) \in B^{\mathcal{C}}_s}}
\phi_{\bm{\mu}_k}^k(\mathbf{c})
}{^{\mathclap{2}}}
\end{equation}
\end{linenomath}
where $S$ denotes the total classes in $\mathcal{O}$. 
Eventually, we use the stochastic gradient descent (SGD) algorithm to update the condensed dataset $\mathcal{C}$ w.r.t the loss $\mathcal{L}$, which can be denoted as
$\mathcal{C} \leftarrow \mathcal{C} - \xi \nabla_{\mathcal{C}} \mathcal{L} $ where $\xi$ is the learning rate. 
We depict the training process in Algorithm \ref{alg:training}. 

Eq. \ref{eq: final_loss} illustrates the many-to-one correspondences between original and condensed samples. Since $\abs{\mathcal{C}} \ll \abs{\mathcal{O}}$, each sample in the condensed batch $B^{\mathcal{C}}_s$ will be exposed to numerous original instances of the same class during training iterations, indicating a mixture of knowledge from multiple cases. This macro-level understanding effectively conceals individual-level information in original data, leading to a \say{bona fide} de-identification. \\

\textbf{Hyper-parameters.} 
Across all three datasets, we apply an Adam \cite{kingma2014adam} optimiser with an initial learning rate of $0.001$ to update $\mathcal{C}$ as operation \ref{op:update_condensed} of Algorithm \ref{alg:training}, and we train every condensed dataset for a total of \numprint{24000} iterations.
The batch size of $B^{\mathcal{O}}_s$ and $B^{\mathcal{C}}_s$ is uniquely set to be 256.
On PhysioNet-2012 and Coswara, we generate three condensed datasets of different samples, i.e., $\abs{\mathcal{C}} =$ 20, 40, and 80, while we set the sizes $\abs{\mathcal{C}}$ of MIMIC-III to be 400, 800, and \numprint{1200}, respectively.
We randomly initialise each condensed dataset from scratch and set the proportions of positive and negative labels to be $50\%$ and $50\%$.
In this work, we set the collection $\Phi$ to consist of three network architectures, namely TCN-\textalpha, ViT-\textalpha, and LSTM-\textalpha. 
We also explore the effects of varying temporal dimensions in condensed samples (see Supplementary Table 9).
\\

\bibliography{ref.bib} 
\bibliographystyle{sn-mathphys}

\end{document}


\title[Supplementary Information]{Supplementary Information - Medical records condensation: a roadmap towards healthcare data democratisation}
\author[1,4]{\fnm{Yujiang} \sur{Wang}}\email{yujiang.wang@oxford-oscar.cn}

\author*[1]{\fnm{Anshul} \sur{Thakur}}\email{anshul.thakur@eng.ox.ac.uk}

\author[2]{\fnm{Mingzhi} \sur{Dong}}\email{mingzhidong@gmail.com} 

\author[3]{\fnm{Pingchuan} \sur{Ma}}\email{pingchuan.ma16@imperial.ac.uk}

\author[3]{\fnm{Stavros} \sur{Petridis}}\email{stavros.petridis04@imperial.ac.uk}

\author*[2]{\fnm{Li} \sur{Shang}}\email{lishang@fudan.edu.cn}

\author[1]{\fnm{Tingting} \sur{Zhu}}\email{tingting.zhu@eng.ox.ac.uk}

\author[1,4]{\fnm{David A.} \sur{Clifton}}\email{david.clifton@eng.ox.ac.uk}

\affil*[1]{\orgdiv{Department of Engineering Science}, \orgname{University of Oxford}, \country{UK}}
\affil*[2]{\orgdiv{School of Computer Science}, \orgname{Fudan University}, \country{China}}

\affil[3]{\orgdiv{Department of Computing}, \orgname{Imperial College London}, \country{UK}}
\affil[4]{\orgdiv{Oxford Suzhou Centre for Advanced Research}, \country{China}}

\makeatletter
\renewcommand{\@seccntformat}[1]{%
  \ifcsname prefix@#1\endcsname
    \csname prefix@#1\endcsname
  \else
    \csname the#1\endcsname\quad
  \fi}
\newcommand\prefix@section{Supplementary Note \thesection~- }
\newcommand{\prefix@subsection}{}
\newcommand{\prefix@subsubsection}{}
\renewcommand{\thesubsection}{\arabic{subsection}}
\renewcommand{\figurename}{Supplementary Figure}
\renewcommand{\tablename}{Supplementary Table}
\newlength\boxwid
\settowidth{\boxwid}{\indent\textbullet\ }
\newcommand{\mdt}{{\boldmath$\cdot$\enspace}}

\makeatother

\maketitle

\section{Datasets}
\subsection{Pre-processing}
\input{tables/supp/dataset_vars/prpr_physionet.tex}
\input{tables/supp/dataset_vars/prpr_mimic3.tex}
\textbf{PhysioNet-2012.} 
The PhysioNet-2012 dataset \cite{silva2012predicting} consists of \numprint{8000} ICU-staying records that are publicly available to study issues of in-hospital mortality. 
A total of 42 variables are presented in each stay, each documented at least once during the first 48 hours after ICU entrance.
37 of those variables are time series with multiple observations during the 48-hour session, while 6 are general descriptors collected once on ICU admissions. 
\say{Weight} is both a general descriptor (measured at admission) and a time series (recorded hourly). 
We convert each time-series variable into a 48-Dimensional vector, each value depicting the state of that hour-session.
We cast each general descriptor into 48-D vectors of constant temporal values (\say{RecordID} are not included).
Notably, we follow the one-hot encoding strategy for the two categorical variables \say{Gender} and \say{ICUType},
with two 48-D binary vectors describing the former and four for the latter.
Two extra scoring variables, SAPS-I \cite{le1984simplified} and SOFA \cite{ferreira2001serial}, are also employed and converted into two 48-channel vectors, respectively.
We stack all vectors from an ICU stay into a representative matrix of shape $48 \times 47$, in which $48$ corresponds to hours and $47$ refers to the features. 
The outcome is a binary variable \say{in-hospital death} indicating whether the patient is a survivor (negative) or died in the hospital (positive), with a total of \numprint{6878} negative samples and \numprint{1122} positive ones, respectively. Supplementary Table \ref{supp_tab:physionet_vars} summarises the clinical variables of PhysioNet-2012. \\

\textbf{MIMIC-III.}
We generally follow \cite{harutyunyan2019multitask} to pre-process the MIMIC-III dataset \cite{johnson2016mimic,johnson2016mimic-physionet}.
In particular,
the MIMIC-III dataset used in this work comprises \numprint{21156} ICU stays of 17 clinical variables selected from tables \say{CHARTEVENTS} and \say{LABEVENTS} (see \cite{johnson2016mimic} for details of those tables).
Those variables are documented during the first 48 hours of ICU admission and can be divided into 5 categorical and 12 continuous. 
Each continuous variable is converted into a 48-D numeric vector, and we encode each categorical variable into one-hot vectors.
All missing values are imputed following \cite{harutyunyan2019multitask}, and we have attached the 17 binary mask vectors indicating whether a variable's values are true or imputed as extra input information.
After pre-processing, we can generate $60$ 48-D vectors representing each ICU stay, i.e., a $48 \times 60$ matrix.
We consider the \say{in-hospital mortality} as the outcome label with 18357 negative (\say{survivor}) and 2799 positive (\say{died in the hospital}) cases. 
Supplementary Table \ref{supp_tab:mimic3_vars} illustrates all clinical variables in MIMIC-III.
\\

\textbf{Coswara.}
The Coswara dataset \cite{sharma2020coswara} consists of \numprint{1368} breathing sounds sampled from participants with polymerase chain reaction (PCR) test results, and the outcome is diagnosing coronavirus-19 (Covid-19).
The duration of an original sound is approximately 3 to 8 seconds at 48-kHz frequency, and we crop a 3-second region of the highest acoustic intensity per sample to unify the time length.
Then we employ a short-term Fourier transform (STFT) with a 100ms FFT window of 50\% overlap to extract the 64 log mel-bands Mel-spectrogram.
This extraction results in a $94 \times 64$ representation matrix for each breathing sound, and $94$ and $64$ are the temporal and acoustic dimensions, respectively.
This dataset contains 948/420 samples with positive/negative PCR results.

\subsection{Dataset splits}
\input{tables/supp/data_split}
Across three datasets, we split the data into train, validation, and test sets. 
The train set is used to learn the condensed data and to train DNNs, while the validation and test is used for evaluating performance of DNNs.
Supplementary Table \ref{supp_tab:data_split} provides an overview of the data splits used in this paper.  
\newpage

\section{Deep Neural Networks}
\label{supp_sec:note1}
\subsection{Network Architectures}
We select a cohort of 11 prevalent DNNs for time-series data to evaluate the deep-learning utility of original and condensed train sets. 
Those 11 DNNs are from a total of five neural network families: multi-scale temporal convolution networks (MS-TCNs) \cite{martinez2020lipreading}, transformers (TRSFs) \cite{vaswani2017attention}, vision transformers (ViTs) \cite{dosovitskiy2021an}, long short term memory networks (LSTMs) \cite{hochreiter1997long}, and recurrent neural networks (RNNs) \cite{hopfield1982neural}. \\

\input{tables/supp/net_architecture/tcn-arch.tex}

\textbf{Multi-scale temporal convolution networks.}
Multi-scale temporal convolution networks (MS-TCNs) \cite{martinez2020lipreading} are a variant of temporal convolution networks (TCNs) \cite{bai2018empirical} and have achieved state-of-the-art performance in various computer vision tasks such as lip-reading. 
TCNs utilise 1D causal convolutions to analyse the temporal information in time-serial data, and MS-TCNs improve their architectures by introducing multi-scale convolutions of varying kernel sizes to better reflect the temporal information of different scales. 
We employ three MS-TCNs in this work, denoted as TCN-\textalpha, TCN-\textbeta, and TCN-\textgamma.
All networks consist of two temporal convolutional (TC) layers, followed by a temporal pooling and a fully connected (FC) layer.
Each TC layer in TCN-\textalpha~comprises three branches of kernel sizes 3, 5, and 7, respectively, and each branch's hidden neuron number is set to 64. 
For TCN-\textbeta, each TC layer only contains a single convolutional branch of kernel size 3 with hidden dimension 64.
Every TC layer in TCN-\textgamma~is set to be a double-branch structure with 64 hidden neurons and a kernel size of 3 and 5 in each branch.
Supplementary Table \ref{supp_tab:tcn_arch} demonstrates the architectures of TCN-\textalpha, TCN-\textbeta, and TCN-\textgamma. \\

\input{tables/supp/net_architecture/trsf-vit-arch.tex}
\textbf{Transformers and vision transformers.}
Transformers (TRSFs) \cite{vaswani2017attention} is initially proposed to address Natural Language Processing (NLP) issues.
The self-attention mechanism enables transformers to capture the input signals' global dependencies; therefore, they are extensively applied in various fields with impressive results.
The original transformer \cite{vaswani2017attention} is constituted of one encoder and one decoder. In this work, we employ the transformer encoder to extract the embedding of time-serial data and discard the decoder. 
We mainly examine two transformer encoders of varying architectures, TRSF-\textalpha~and TRSF-\textbeta.
Both networks comprise two multi-headed self-attention (MSA) layers followed by a temporal pooling and a multi-layer perceptron (MLP) head of two FC layers. 
We ignore the positional embedding for simplicity.
In TRSF-\textalpha, we set each MSA layer to include 16 self-attention heads, and each head consists of 64 attention channels. The hidden neurons in MLP are set to 64.
Each MSA layer in TRSF-\textbeta~comprises 4 attention heads of 256 channels, and the hidden dimension of its MLP is 128.

Vision transformers (ViTs) \cite{dosovitskiy2021an} are a variant of transformers designed for image recognition. 
Dividing each image into patches and linearly projecting patches into sequential embedding, ViTs can be used to analyse images with state-of-the-art performance. 
ViTs also differ from transformers in obtaining the sequence representation. A learnable token is pre-appended to the input and its corresponding part at the output side is fed into the MLP. 
We discard the linear projection component in this work and feed time-serial data directly into the ViT encoder.
A total of two ViTs with different architectures are inspected, ViT-\textalpha~and ViT-\textbeta.
Both ViTs contain four MSA layers, and the corresponding output of the learnable token is fed into an MLP with two FC layers, while the positional embedding is also omitted for simplicity.
ViT-\textalpha~possesses 16 self-attention heads with 64 channels per MSA layer, while each MSA of 
 ViT-\textbeta~is made up of 4 heads of 128 attention channels.
 Supplementary Table \ref{supp_tab:trfs_vit_arch} illustrates the architectures of TRSF-\textalpha, TRSF-\textbeta, ViT-\textalpha, and ViT-\textbeta~in details. \\

\textbf{LSTMs and RNNs.}
\input{tables/supp/net_architecture/lstm-rnn-arch.tex}
Long short term memory networks (LSTMs) \cite{hochreiter1997long} and recurrent neural networks (RNNs) \cite{hopfield1982neural} are both recurrent-based networks but with varying implementations.
RNNs are known to be unable to capture the long-term temporal dependency within data due to their naive architectures, while LSTMs improve this limitation by introducing more sophisticated gating mechanisms. 
In this work, we apply two LSTMs and two RNNs denoted as LSTM-\textalpha~and LSTM-\textbeta~, RNN-\textalpha, and RNN-\textbeta, respectively.
Across the four architectures, the time-series data is fed into one recurrent layer, either LSTM-based or RNN-based, followed by an FC layer for final predictions.
LSTM-\textalpha~differs from LSTM-\textbeta~in that the recurrent layer of the former consists of a hidden state of 256 units, while the latter comprises 128 instead. 
We set the dimension of the hidden state in RNN-\textalpha's recurrent layer to be 256, and 128 for RNN-\textbeta.
Supplementary Table \ref{supp_tab:lstm_rnn_arch} explain the architectures of LSTM-\textalpha~and LSTM-\textbeta, RNN-\textalpha, and RNN-\textbeta.
\newpage

\subsection{Training settings}
\input{tables/supp/train_setting.tex}
Across all three datasets, we trained all 11 networks on the original train set and its condensed versions to evaluate the deep-learning utility preserved by DC.
The binary cross entropy (BCE) loss is employed to train all DNNs, and we apply Adam \cite{kingma2014adam} optimiser for all experiments with varying initial learning rates on different datasets.
We also perform a data standardisation (subtracted by mean and divided by STD) for numeric input data before feeding it into DNNs.
The detailed training hyper-parameters are summarised in Supplementary Table \ref{supp_tab:train_setting}.
Note that those settings should be distinguished from the hyper-parameters of learning condensed data described in Section \say{Methods} of the main manuscript.

\newpage

\section{Results}
\subsection{AUCs of 11 DNNs}
\input{tables/supp/results/pre_dnn_res.tex}

\subsection{Effects of changing temporal dimension in DC}
\input{tables/supp/results/time_dim_effect.tex}
Across the manuscript, we pre-define each condensed sample to retain the size of the original one, i.e., for original instances $\mathbf{x} \in \mathbb{R}^{T \times F}$ where $T$ and $F$ correspond to the temporal and feature dimension, we will learn condensed samples of the same shape $\mathbf{c} \in \mathbb{R}^{T \times F}$. 
In principle, however, we can set the condensed sample to have inconsistent sizes as the original insofar as the embedding distributions between them can still be matched in DC.
In this work, we employ 11 DNNs as the embedding function, while all architectures can accept input with an arbitrary temporal dimension. 
That is, we can define condensed samples $\mathbf{c} \in \mathbb{R}^{T^{\ast} \times F}$ where $T^{\ast} \neq T$ or $T^{\ast} = T$.

We investigate the effects of different temporal dimensions (TDs) in condensed samples in Supplementary Table \ref{supp_tab:time_dim_effects}. 
We can discover that using a smaller $T^{\ast}$ can further compress the data volumes without significantly affecting the test AUCs.
This observation reflects an extra advantage of DC: the flexibility in size control.
After choosing $\abs{\mathcal{C}} \ll \abs{\mathcal{O}}$, we could further compress the sizes of condensed data by picking a $T^{\ast} \lt T$ at the cost of slightly decreased accuracy.

\newpage

\subsection{Learning Curves}
\begin{figure}[h!]
\vspace{-2em}
    \centering
    \caption{
    Learning curves of different networks trained on the original and condensed data across 3 datasets. The red-dotted line indicates the step of convergence. 
    \textbf{a}, RNN-\textbeta~and TRSF-\textbeta~on PysioNet-2012. 
     \textbf{b}, RNN-\textalpha~and TCN-\textbeta~on MIMIC-III. 
          \textbf{c}. LSTM-\textalpha~and TCN-\textgamma~on Coswara. 
          The convergence rates of DNNs trained on the condensed data significantly outperform the same networks trained on the original. 
    } 
    \includegraphics[width=0.95\textwidth]{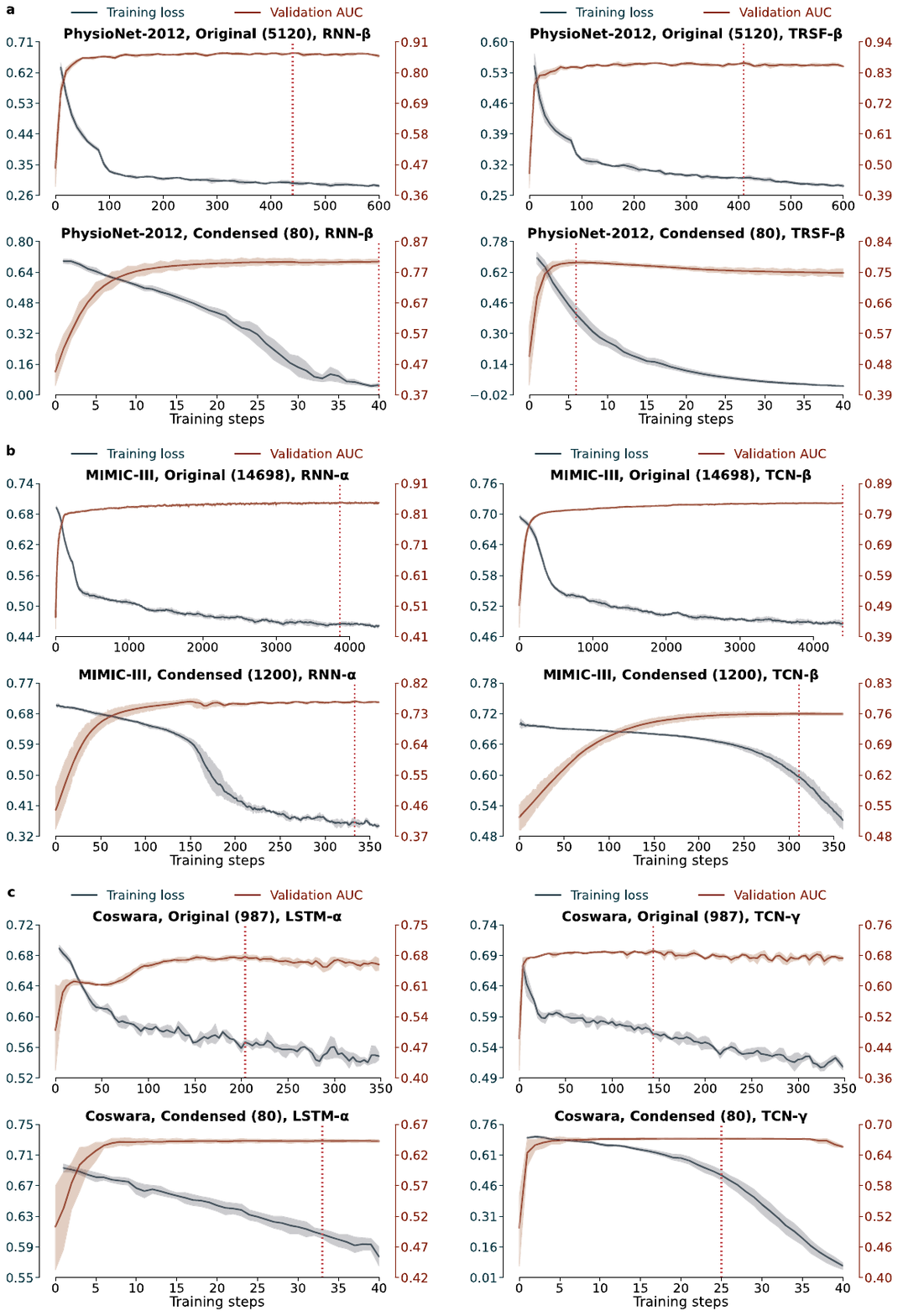}
    \label{supp_fig: learning_curve_extra}
\end{figure} 
\newpage

\subsection{Learning time of DC}
\input{tables/supp/time_costs.tex}

\newpage
\subsection{Data Visualisation}
\begin{figure}[h!]
\vspace{-2em}
    \centering
    \caption{
\textbf{a}, the heatmaps of
three original and three condensed samples randomly picked from PhysioNet. 
\textbf{b}, the mel-spectrogram plots of three random audio segments and three random condensed samples on Coswara.
Condensed instances behave like random noises without regular patterns in the original ones.
    } 
    \includegraphics[width=1.0\textwidth]{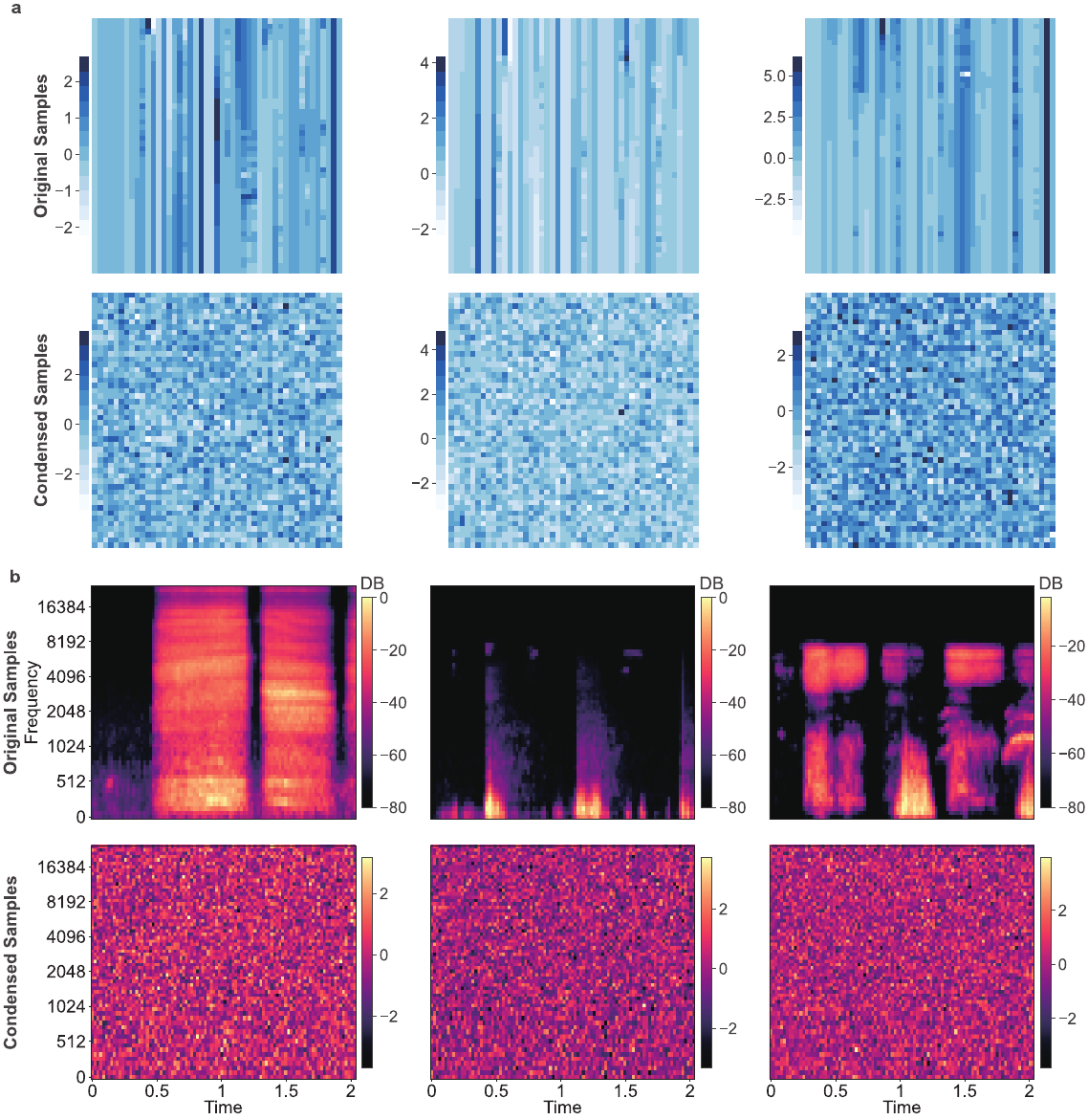}
    \label{supp_fig: data_heatmap_extra}
\end{figure}
\newpage
\begin{figure}[h!]
    \centering
    \caption{
The 48-hour trends of 12 additional clinical variables from PhysioNet-2012 computed from 80 original and 80 condensed instances, respectively.
Those variables are: Alanine transaminase (\protect\say{ALT}), Albumin, Blood urea nitrogen (\protect\say{BUN}), Bilirubin, Serum creatinine (\protect\say{Creatinine}), Troponin-I, Glasgow Coma Score (\protect\say{GCS}), Temperature (\protect\say{Temp}),
Serum glucose (\protect\say{Glucose}),
Hematocrit (\protect\say{HCT}), 
Serum potassium (\protect\say{K}), and Serum magnesium (\protect\say{Mg}).
    } 
    \includegraphics[width=1.0\textwidth]{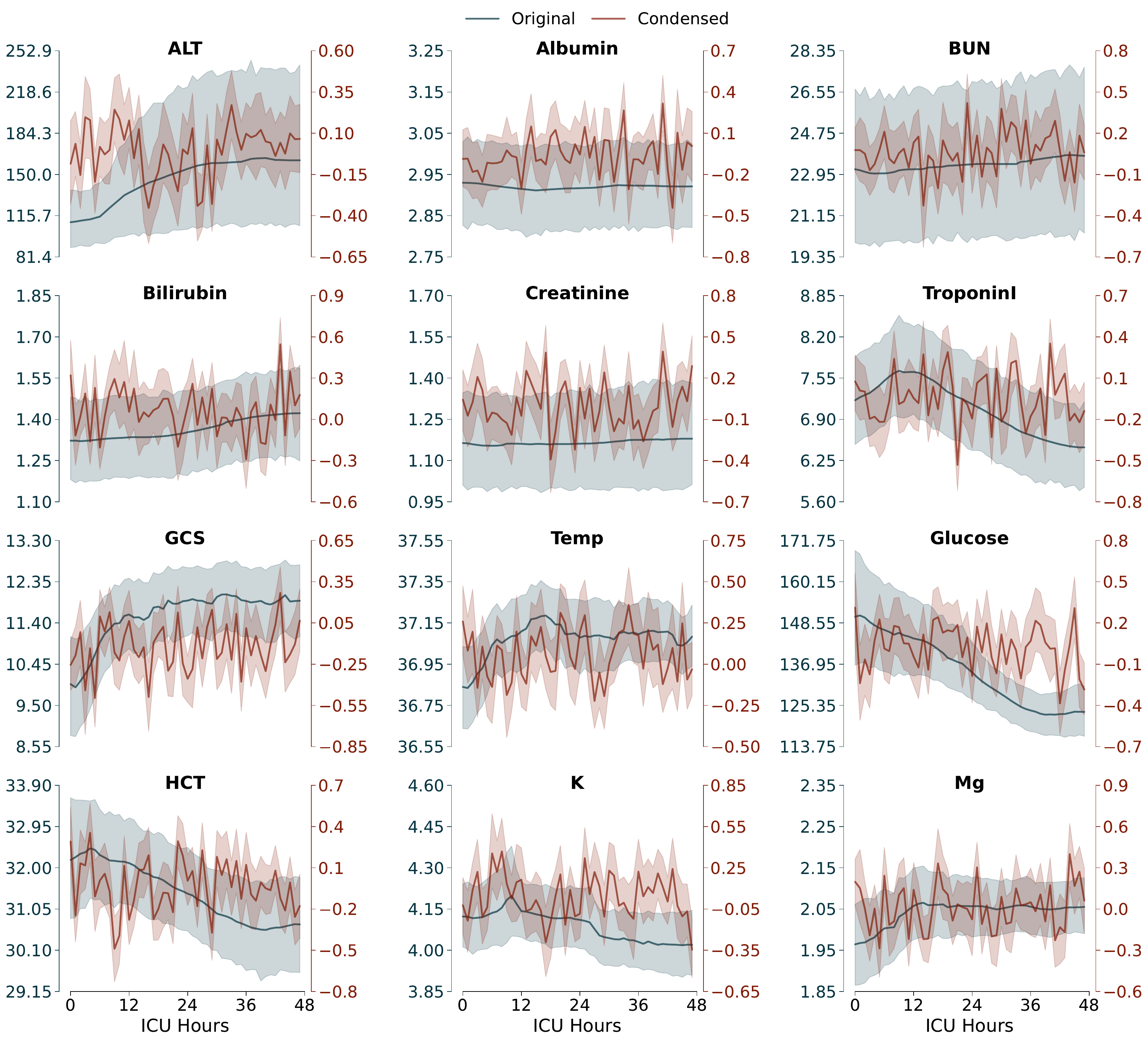}
    \label{supp_fig: var_physio_extra}
\end{figure}
\newpage
\begin{figure}[h]
    \centering
    \caption{
The 3D distributions of 987 original and 80 condensed samples embedded into 3D with t-SNE on Coswara.
    } 
    \includegraphics[width=0.67\textwidth]{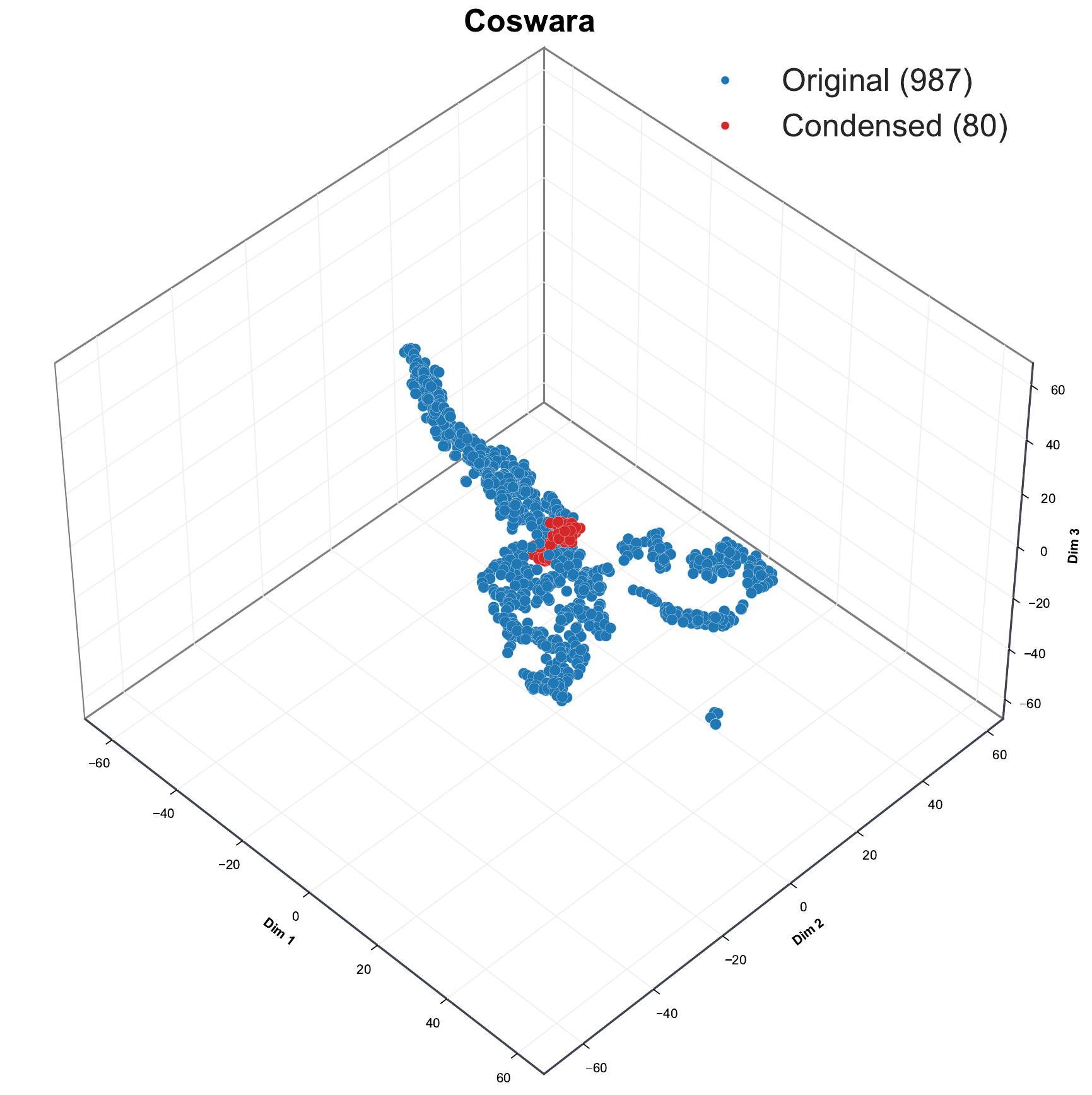}
    \label{supp_fig: tsne_coswara}
\end{figure}

\bibliography{ref.bib} 
\bibliographystyle{sn-mathphys}


%% file: tables/sumup.tex
\begin{table}[t!]
\centering
\caption{The average performance of DNNs learned on the original train set and three condensed groups on PhysioNet-2012, MIMIC-III, and Coswara datasets. We repeat the training of each DNN five times and compute the mean and standard deviation (SD) of all 11 DNNs. For each data collection, we also report its space consumption in megabytes (MBs).}
\label{tab:sum_up}
\begin{tabularx}{1.0\textwidth}{ykjj}
\toprule
\textbf{Datasets} & \textbf{Train data} (samples) & \textbf{Sizes} (MBs) & \textbf{Test AUC} (SD) \\
\cmidrule(lr){2-2} 
\cmidrule(lr){3-4} 
\multirow{4}{*}{PhysioNet-2012} & Original (5120)  & 88.13     & 0.858 (0.011) \\
 & Condensed (80)       & 0.69      & 0.804 (0.014) \\
 & Condensed (40)       & 0.34      & 0.804 (0.016) \\
 & Condensed (20)       & 0.17      & 0.803 (0.012) \\
 \cmidrule(lr){2-2} 
\cmidrule(lr){3-4} 
\multirow{4}{*}{MIMIC-III}      & Original (14698)     & 322.95    & 0.840 (0.007) \\
 & Condensed (1200)     & 13.18     & 0.756 (0.014) \\
 & Condensed (800)      & 8.79      & 0.750 (0.014) \\
 & Condensed (400)      & 4.39      & 0.741 (0.019) \\
\cmidrule(lr){2-2} 
\cmidrule(lr){3-4} 
\multirow{4}{*}{Coswara}        & Original (987)       & 46.27     & 0.737 (0.017) \\
 & Condensed (80)       & 1.88      & 0.642 (0.016) \\
 & Condensed (40)       & 0.94      & 0.641 (0.021) \\
 & Condensed (20)       & 0.47      & 0.632 (0.030) \\
 \bottomrule
\end{tabularx}
\end{table}

%% file: tables/training_steps.tex
\begin{table}[ht]
\centering
\vspace{-2em}
\caption{The training steps of convergence across 11 DNNs trained on the original and condensed data. \protect\say{Ori}. and\protect\say{Con.} refers to \protect\say{Originial} and \protect\say{Condensed}, respectively, and \protect\say{AVG} is the average steps. We post-pend the number of samples to each data group. }
\label{tab:training_steps}
\tiny
\begin{tabularx}{1.0\textwidth}{yjtiiiiiiiiiii}
\toprule
\multirow{4}{*}{\textbf{Dataset}} & \multirow{4}{*}{\textbf{\shortstack{Train\\data}}} & \multicolumn{12}{c}{\textbf{Training Steps of Convergence}}  \\  
\cmidrule(l){3-14} 
&  & AVG & TCN-\textalpha & TCN-\textbeta & TCN-\textgamma & ViT-\textalpha & ViT-\textbeta & TRSF-\textalpha & TRSF-\textbeta & LSTM-\textalpha & LSTM-\textbeta & RNN-\textalpha & RNN2-\textbeta \\
 \cmidrule(lr){1-2}  \cmidrule(lr){3-3}  \cmidrule(l){4-14} 
\multirow{2}{*}{\shortstack{PhysioNet\\-2012}} & Ori.(5120) & 330 & 240 & 370 & 190 & 500 & 340 & 200  & 410  & 170  & 300  & 480 & 440 \\ 
& Con.(80) & 15  & 7 & 19 & 5 & 2 & 3 & 7 & 6 & 16 & 27 & 33 & 40 \\
 \cmidrule(lr){2-2}  \cmidrule(lr){3-3}  \cmidrule(l){4-14} 
\multirow{2}{*}{MIMIC-III} & Ori.(14698) & 3083  & 4310 & 4400 & 3960 & 2030 & 1340 & 1860 & 2240   & 2500   & 3020 & 3870  & 4390 \\ 
& Con.(1200) & 273 & 356 & 311 & 361 & 32 & 358  & 81 & 361 & 186 & 323 & 333 & 304 \\
 \cmidrule(lr){2-2}  \cmidrule(lr){3-3}  \cmidrule(l){4-14} 
\multirow{2}{*}{Coswara}  & Ori.(987) & 217 & 148 & 384 & 144 & 132 & 108  & 164 & 104  & 204  & 324  & 324 & 360 \\ 
& Con.(80) & 26 & 3 & 40 & 25 & 25 & 25 & 40 & 32 & 33 & 24  & 5 & 35 \\
\bottomrule
\end{tabularx}
\end{table}

%% file: tables/learning_algorithm.tex
\begin{algorithm}[t!]
\caption{Learning a condensed healthcare dataset}
\label{alg:training}
\textbf{Input:} The original healthcare dataset $\mathcal{O}$  \\
\textbf{Require:} The randomly-initialised condensed dataset $\mathcal{C}$ with $\abs{\mathcal{C}}$ samples of $S$ classes, the collection of deep networks $\Phi$, the parameter distribution $P_{\bm{\mu}_k}$ for arbitrary network in $\Phi$, the training iterations $M$, the learning rate $\xi$
\begin{algorithmic}[1]
\For{$m=1:M$}
\State Sample network $\phi_{\bm{\mu}_k}^k \sim \Phi$
\State Sample network parameters $\bm{\mu}_k \sim P_{\bm{\mu}_k}$
\For{$s=1:S$}
\State Sample an original batch $B^{\mathcal{O}}_s \sim \mathcal{O}$ 
\State Sample a condensed batch $B^{\mathcal{C}}_s \sim \mathcal{C}$
\EndFor
\State Compute loss $\mathcal{L}$ as in Eq. \ref{eq: final_loss}
\State Update $\mathcal{C} \leftarrow \mathcal{C} - \xi \nabla_{\mathcal{C}} \mathcal{L}$  \label{op:update_condensed}
\EndFor 
\end{algorithmic}
\textbf{Output:} The condensed healthcare dataset $\mathcal{C}$
\end{algorithm}

%% file: tables/supp/dataset_vars/prpr_physionet.tex
\begin{table}[]
\caption{The clinical variables in PhysioNet-2012. 
We create one or multiple 48-D vectors for each variable.
We apply one-hot encoding to represent \protect\say{Gender} and \protect\say{ICUType}, and we specify the categorical value with \protect\say{\textrightarrow}.
\protect\say{Counts(Vec.)} refer to the variable count post-pending number of vectors after pre-processing.
Each ICU stay leads to a matrix of size $48 \times 47$ with \protect\say{in-hospital death} as the outcome label.
}
\centering
\label{supp_tab:physionet_vars}
\footnotesize
\begin{tabularx}{1.0\textwidth}{>{\raggedright\arraybackslash \hangindent=0.9em}k>{\raggedright\arraybackslash \hangindent=0.9em}k>{\raggedright\arraybackslash}t>{\centering\arraybackslash}y}
\toprule
\multicolumn{2}{c}{\textbf{Clinical Variables}} & {\textbf{Type}} & \textbf{Count}(Vec.) \\
 \cmidrule(lr){1-2} \cmidrule(lr){3-3} \cmidrule(l){4-4} 
\mdt{Albumin} & \mdt{Alkaline phosphatase} & \multirow{19}{*}{\shortstack{Time\\series}} & \multirow{19}{*}{37(37)}  \\
\mdt{Blood urea nitrogen} & \mdt{Cholesterol} &  &    \\
\mdt{Glasgow Coma Score} & \mdt{Serum glucose} &  &    \\
\mdt{Serum potassium} & \mdt{Lactate} &  &    \\
\mdt{Serum sodium} & \mdt Urine output  &  &    \\
\mdt{Serum creatinine}  & \mdt{Temperature} &  &    \\
\mdt{Bilirubin} & \mdt{Fractional inspired O2} &  &    \\
\mdt{Mechanical ventilation respiration}  & \mdt{Partial pressure of arterial CO2} &  &    \\
\mdt{Alanine transaminase} & \mdt{Aspartate transaminase} &  &    \\
\mdt Invasive systolic arterial blood pressure & \mdt Invasive diastolic arterial blood pressure &  &    \\
\mdt{Non-invasive diastolic arterial blood pressure}  & \mdt Non-invasive mean arterial blood pressure &  &    \\
\mdt Serum bicarbonate & \mdt Hematocrit &  &    \\
\mdt Non-invasive systolic arterial blood pressure & \mdt Invasive mean arterial blood pressure &  &    \\
\mdt Arterial pH & \mdt Serum magnesium &  &    \\
\mdt Platelets & \mdt Respiration rate &  &    \\
\mdt Troponin-I & \mdt Troponin-T &  &    \\
\mdt Heart rate & \mdt White blood cell count &  &    \\
\mdt O2 saturation in hemoglobin & \mdt{Partial pressure of arterial O2} &  &    \\
\mdt Weight &  &    \\
 \cmidrule(lr){1-2} 
\mdt Age & \mdt Gender\textrightarrow Female &  &    \\
\mdt Height &\mdt Gender\textrightarrow Male &  &    \\
\mdt ICUType\textrightarrow Surgical ICU  & \mdt ICUType\textrightarrow Medical ICU &  &    \\
\mdt ICUType\textrightarrow Cardiac Surgery Recovery Unit & \mdt ICUType\textrightarrow Coronary Care Unit & \multirow{-4}{*}{\shortstack[l]{General\\descriptor}} & \multirow{-4}{*}{4(8)} \\
\cmidrule(lr){1-2} 
\mdt SAPS-I & \mdt SOFA & Scoring & 2(2) \\
\cmidrule(lr){1-2} \cmidrule(lr){3-3}  \cmidrule(lr){4-4}
\multicolumn{2}{l}{\mdt In-Hospital Death} & Outcome & 1(1)
 \\
 \bottomrule
 \end{tabularx}
\end{table}

%% file: tables/supp/dataset_vars/prpr_mimic3.tex
\begin{table}[]
\caption{The clinical variables in MIMIC-III. 
For each variable, we create one or multiple 48-D vectors with imputed values following \cite{harutyunyan2019multitask}.
We apply one-hot encoding to represent each categorical variable, and we detail the categorical value using \protect\say{\textrightarrow}.
Those imputation-indicating mask vectors are seen as inputs. 
\protect\say{GCS} refers to \protect\say{Glasgow coma scale}, while \protect\say{verb. resp.} is \protect\say{verbal response}.
Each ICU stay leads to a $48 \times60$ matrix with \protect\say{in-hospital death} as the outcome.
}
\centering
\label{supp_tab:mimic3_vars}
\footnotesize
\begin{tabularx}{1.0\textwidth}{>{\raggedright\arraybackslash \hangindent=0.9em}s>{\raggedright\arraybackslash \hangindent=0.9em}s>{\raggedright\arraybackslash}t>{\centering\arraybackslash}g}
\toprule
\multicolumn{2}{c}{\textbf{Clinical Variables}} & \textbf{Type} & \textbf{Count}(Vec.) \\
 \cmidrule(lr){1-2} \cmidrule(lr){3-3} \cmidrule(l){4-4} 
\mdt Diastolic blood pressure & \mdt Oxygen saturation & \multirow{6}{*}{\shortstack{Conti-\\nuous}} & \multirow{6}{*}{12(12)} \\
\mdt Fraction inspired oxygen & \mdt Respiratory rate &  &  \\
\mdt Glucose & \mdt Systolic blood pressure &  &  \\
\mdt Heart Rate & \mdt Temperature &  &  \\
\mdt Height & \mdt Weight &  &  \\
\mdt Mean blood pressure & \mdt pH &  &  \\
 \cmidrule(lr){1-2} 
\mdt Capillary refill rate \textrightarrow~0 & \mdt Capillary refill rate \textrightarrow~1 & \multirow{16}{*}{\shortstack{Categ-\\orical}} & \multirow{16}{*}{5(31)} \\
\mdt GCS eye opening \textrightarrow~To speech & \mdt GCS eye opening \textrightarrow~None&  &  \\
\mdt GCS eye opening \textrightarrow~To Pain & \mdt GCS verb. resp. \textrightarrow~Confused  &  &  \\\mdt GCS eye opening \textrightarrow~Spontaneously  & \mdt GCS eye opening \textrightarrow~No Response &  &  \\
\mdt GCS verb. resp. \textrightarrow~Incomprehensible sounds & \mdt GCS verb. resp. \textrightarrow~Inappropriate Words &  &  \\
\mdt GCS verb. resp. \textrightarrow~No Response  & \mdt GCS verb. resp. \textrightarrow~Oriented &  &  \\
\mdt GCS motor response \textrightarrow~Abnormal extension &  \mdt GCS motor response \textrightarrow~Abnorm flexion  &  &  \\
\mdt GCS motor response \textrightarrow~No Response & \mdt GCS motor response \textrightarrow~Flex-withdraws &  &  \\
\mdt GCS motor response \textrightarrow~Localizes Pain & \mdt GCS motor response \textrightarrow~Obeys Commands &  &  \\
\mdt GCS total \textrightarrow~3 & \mdt GCS total \textrightarrow~10 &  &  \\
\mdt GCS total \textrightarrow~4 & \mdt GCS total \textrightarrow~11 &  &  \\
\mdt GCS total \textrightarrow~5 & \mdt GCS total \textrightarrow~12 &  &  \\
\mdt GCS total \textrightarrow~6 & \mdt GCS total \textrightarrow~13 &  &  \\
\mdt GCS total \textrightarrow~7 & \mdt GCS total \textrightarrow~14 &  &  \\
\mdt GCS total \textrightarrow~8 & \mdt GCS total \textrightarrow~15 &  &  \\
\mdt GCS total \textrightarrow~9 &  &  &  \\
 \cmidrule(lr){1-2} 
\mdt mask \textrightarrow~Capillary refill rate & \mdt mask \textrightarrow~GCS eye opening & \multirow{9}{*}{\shortstack{Mask}} & \multirow{9}{*}{17(17)} \\
\mdt mask \textrightarrow~Fraction inspired oxygen & \mdt mask \textrightarrow~Diastolic blood pressure  &  &  \\
\mdt mask \textrightarrow~GCS motor response & \mdt mask \textrightarrow~GCS verb. resp. &  &  \\
\mdt mask \textrightarrow~GCS total & \mdt mask \textrightarrow~Glucose &  &  \\
\mdt mask \textrightarrow~Heart Rate & \mdt mask \textrightarrow~Height &  &  \\
\mdt mask \textrightarrow~Mean blood pressure & \mdt mask \textrightarrow~Oxygen saturation &  &  \\
\mdt mask \textrightarrow~Respiratory rate & \mdt mask \textrightarrow~Systolic blood pressure &  &  \\
\mdt mask \textrightarrow~Temperature & \mdt mask \textrightarrow~Weigh &  &  \\
\mdt mask \textrightarrow~pH &  &  &  \\
 \cmidrule(lr){1-2}  \cmidrule(lr){3-3}  \cmidrule(lr){4-4} 
\mdt In-Hospital Death &  & Outcome & 1(1)
 \\
 \bottomrule
 \end{tabularx}
\end{table}

%% file: tables/supp/data_split.tex
\begin{table}[h!]
\vspace{-2em}
\caption{Number of samples (number of positive/negative instances) in the train, validation, and test sets across three datasets.
}
\centering
\label{supp_tab:data_split}
\footnotesize
\begin{tabularx}{1.0\textwidth}{kskk}
\toprule
\textbf{Dataset} & \textbf{Train} & \textbf{Validation} & \textbf{Test} \\ \cmidrule(lr){1-1}  \cmidrule(lr){2-4} 
PhysioNet-2012 & \numprint{5120} (728/\numprint{4392}) & \numprint{1280} (167/\numprint{1113}) & \numprint{1600} (227/\numprint{1373}) \\
MIMIC-III & \numprint{14698} (\numprint{1989}/\numprint{12709}) & \numprint{3222} (436/\numprint{2786}) & \numprint{3236} (374/\numprint{2862}) \\
Coswara & 987 (683/304) & 175 (121/54) & 206 (144/62)
 \\
 \bottomrule
 \end{tabularx}
\end{table}

%% file: tables/supp/net_architecture/tcn-arch.tex
\begin{table}[]
\centering
\caption{The architectures of TCN-\textalpha, TCN-\textbeta, and TCN-\textgamma. 
The kernel size is post-pended to each \textit{Conv1D} operation, and the dropout rate is also indicated following each \textit{Dropout}.
We apply PReLU \cite{he2015delving} as the main activation function.
\protect\say{RC$i$\textrightarrow} refers to the starting of the $i$-th residual connection, and \protect\say{RC$i$\textleftarrow} for the ending.
\protect\say{Concat.} stands for channel-wise concatenation, while \protect\say{Chmop1D} is the causal convolution implementation \cite{bai2018empirical}. 
}
\label{supp_tab:tcn_arch}
\footnotesize
\begin{tabularx}{1.0\textwidth}{iyyyyyy}
\toprule
\multirow{2}{*}{Layer} & \multicolumn{3}{c}{TCN-\textalpha} & \multirow{2}{*}{TCN-\textbeta} & \multicolumn{2}{c}{TCN-\textgamma} \\
 & Branch1 & Branch2 & Branch3 &  & Branch1 & Branch2 \\
  \cmidrule(lr){1-1}  \cmidrule(lr){2-4}  \cmidrule(lr){5-5}  \cmidrule(lr){6-7}
\multirow{14}{*}{1} & \multicolumn{3}{c}{RC1\textrightarrow} & RC1\textrightarrow & \multicolumn{2}{c}{RC1\textrightarrow} \\
  \cmidrule(lr){2-2}  \cmidrule(lr){3-3}  \cmidrule(lr){4-4} \cmidrule(lr){6-6}  \cmidrule(lr){7-7}  
 & Conv1D(3) & Conv1D(5) & Conv1D(7) & Conv1D(3) & Conv1D(3) & Conv1D(5) \\
 & BatchNorm & BatchNorm & BatchNorm & BatchNorm & BatchNorm & BatchNorm \\
 & Chomp1D & Chomp1D & Chomp1D & Chomp1D & Chomp1D & Chomp1D \\
 & PReLU & PReLU & PReLU & PReLU & PReLU & PReLU \\
  \cmidrule(lr){2-2}  \cmidrule(lr){3-3}  \cmidrule(lr){4-4} \cmidrule(lr){6-6}  \cmidrule(lr){7-7}  
 & \multicolumn{3}{c}{Concat. \& Dropout(0.5)} & Dropout(0.5) & \multicolumn{2}{c}{Concat. \& Dropout(0.5)} \\
  \cmidrule(lr){2-2}  \cmidrule(lr){3-3}  \cmidrule(lr){4-4} \cmidrule(lr){6-6}  \cmidrule(lr){7-7}  
 & Conv1D(3) & Conv1D(5) & Conv1D(7) & Conv1D(3) & Conv1D(3) & Conv1D(5) \\
 & BatchNorm & BatchNorm & BatchNorm & BatchNorm & BatchNorm & BatchNorm \\
 & Chomp1D & Chomp1D & Chomp1D & Chomp1D & Chomp1D & Chomp1D \\
 & PReLU & PReLU & PReLU & PReLU & PReLU & PReLU \\
   \cmidrule(lr){2-2}  \cmidrule(lr){3-3}  \cmidrule(lr){4-4} \cmidrule(lr){6-6}  \cmidrule(lr){7-7}  
 & \multicolumn{3}{c}{Concat. \& Dropout(0.5)} & Dropout(0.5) & \multicolumn{2}{c}{Concat. \& Dropout(0.5)} \\
 & \multicolumn{3}{c}{Conv1D(1)} & Conv1D(1) & \multicolumn{2}{c}{Conv1D(1)} \\
 & \multicolumn{3}{c}{RC1\textleftarrow} & RC1\textleftarrow & \multicolumn{2}{c}{RC1\textleftarrow} \\
 & \multicolumn{3}{c}{PReLU} & PReLU & \multicolumn{2}{c}{PReLU} \\
 \cmidrule(lr){2-4}  \cmidrule(lr){5-5}  \cmidrule(lr){6-7}
\multirow{14}{*}{2} & \multicolumn{3}{c}{RC2\textrightarrow} & RC2\textrightarrow & \multicolumn{2}{c}{RC2\textrightarrow} \\
   \cmidrule(lr){2-2}  \cmidrule(lr){3-3}  \cmidrule(lr){4-4} \cmidrule(lr){6-6}  \cmidrule(lr){7-7}  
 & Conv1D(3) & Conv1D(5) & Conv1D(7) & Conv1D(3) & Conv1D(3) & Conv1D(5) \\
 & BatchNorm & BatchNorm & BatchNorm & BatchNorm & BatchNorm & BatchNorm \\
 & Chomp1D & Chomp1D & Chomp1D & Chomp1D & Chomp1D & Chomp1D \\
 & PReLU & PReLU & PReLU & PReLU & PReLU & PReLU \\
    \cmidrule(lr){2-2}  \cmidrule(lr){3-3}  \cmidrule(lr){4-4} \cmidrule(lr){6-6}  \cmidrule(lr){7-7}  
 & \multicolumn{3}{c}{Concat. \& Dropout(0.5)} & Dropout(0.5) & \multicolumn{2}{c}{Concat. \& Dropout(0.5)} \\
 & Conv1D(3) & Conv1D(5) & Conv1D(7) & Conv1D(3) & Conv1D(3) & Conv1D(5) \\
 & BatchNorm & BatchNorm & BatchNorm & BatchNorm & BatchNorm & BatchNorm \\
 & Chomp1D & Chomp1D & Chomp1D & Chomp1D & Chomp1D & Chomp1D \\
 & PReLU & PReLU & PReLU & PReLU & PReLU & PReLU \\
    \cmidrule(lr){2-2}  \cmidrule(lr){3-3}  \cmidrule(lr){4-4} \cmidrule(lr){6-6}  \cmidrule(lr){7-7}  
 & \multicolumn{3}{c}{Concat. \& Dropout(0.5)} & Dropout(0.5) & \multicolumn{2}{c}{Concat. \& Dropout(0.5)} \\
 & \multicolumn{3}{c}{Conv1D(1)} & Conv1D(1) & \multicolumn{2}{c}{Conv1D(1)} \\
 & \multicolumn{3}{c}{RC2\textleftarrow} & RC2\textleftarrow & \multicolumn{2}{c}{RC2\textleftarrow} \\
 & \multicolumn{3}{c}{PReLU} & PReLU & \multicolumn{2}{c}{PReLU} \\
  \cmidrule(lr){2-4}  \cmidrule(lr){5-5}  \cmidrule(lr){6-7}
\multirow{3}{*}{3} & \multicolumn{3}{c}{Pooling} & Pooling & \multicolumn{2}{c}{Pooling} \\
& \multicolumn{3}{c}{Linear} & Linear & \multicolumn{2}{c}{Linear} \\
 & \multicolumn{3}{c}{Sigmoid} & Sigmoid & \multicolumn{2}{c}{Sigmoid} \\
 \bottomrule
 \end{tabularx}
\end{table}

%% file: tables/supp/net_architecture/trsf-vit-arch.tex
\begin{table}[]
\centering
\caption{The architectures of TRSF-\textalpha, TRSF-\textbeta, ViT-\textalpha, and ViT-\textbeta. 
\protect\say{MSA} refers to the Multi-head Self-attention Layer, and the head number and attention channel are post-pended.
\protect\say{\protect\verb|[token]|\textleftarrow} refers to appending a learning token to input, and \protect\say{\protect\verb|[token]|\textrightarrow} means getting the token at output side.
\protect\say{F.F.} stands for the feed-forward layer and consists of two FC layers.
We apply GELU \cite{hendrycks2016gaussian} as the major activation function.
}
\label{supp_tab:trfs_vit_arch}
\footnotesize
\begin{tabularx}{1.0\textwidth}{ijjjj}
\toprule
 Layer & TRSF-\textalpha & TRSF-\textbeta & ViT-\textalpha & ViT-\textbeta \\
\cmidrule(lr){1-1} \cmidrule(lr){2-2} \cmidrule(lr){3-3}   \cmidrule(lr){4-4}
\cmidrule(lr){5-5}
\multirow{9}{*}{1} & \multirow{2}{*}{RC1a\textrightarrow} & \multirow{2}{*}{RC1a\textrightarrow} & \verb|[token]|\textleftarrow & \verb|[token]|\textleftarrow \\
 &  &  & RC1a\textrightarrow & RC1a\textrightarrow \\
 & LayerNorm & LayerNorm & LayerNorm & LayerNorm \\
 & MSA(16,64) & MSA(4,256) & MSA(16,64) & MSA(4,256) \\
 & RC1a\textleftarrow & RC1a\textleftarrow & RC1a\textleftarrow & RC1a\textleftarrow \\
 & RC1b\textrightarrow & RC1b\textrightarrow & RC1b\textrightarrow & RC1b\textrightarrow \\
 & LayerNorm & LayerNorm & LayerNorm & LayerNorm \\
 & F.F. & F.F. & F.F. & F.F. \\
 & RC1b\textleftarrow & RC1b\textleftarrow & RC1b\textleftarrow & RC1b\textleftarrow \\
 \cmidrule(lr){2-2} \cmidrule(lr){3-3}   \cmidrule(lr){4-4}
\cmidrule(lr){5-5}
\multirow{8}{*}{2} & RC2a\textrightarrow & RC2a\textrightarrow & RC2a\textrightarrow & RC2a\textrightarrow \\
 & LayerNorm & LayerNorm & LayerNorm & LayerNorm \\
 & MSA(16,64) & MSA(4,256) & MSA(16,64) & MSA(4,256) \\
 & RC2a\textleftarrow & RC2a\textleftarrow & RC2a\textleftarrow & RC2a\textleftarrow \\
 & RC2b\textrightarrow & RC2b\textrightarrow & RC2b\textrightarrow & RC2b\textrightarrow \\
 & LayerNorm & LayerNorm & LayerNorm & LayerNorm \\
 & F.F. & F.F. & F.F. & F.F. \\
 & RC2b\textleftarrow & RC2b\textleftarrow & RC2b\textleftarrow & RC2b\textleftarrow \\
 \cmidrule(lr){2-2} \cmidrule(lr){3-3}   \cmidrule(lr){4-4}
\cmidrule(lr){5-5}
\multirow{8}{*}{3} & Pooling & Pooling & RC3a\textrightarrow & RC3a\textrightarrow \\
 & LayerNorm & LayerNorm & LayerNorm & LayerNorm \\
 & Linear & Linear & MSA(16,64) & MSA(4,256) \\
 & GELU & GELU & RC3a\textleftarrow & RC3a\textleftarrow \\
 & Droput(0.5) & Droput(0.5) & RC3b\textrightarrow & RC3b\textrightarrow \\
 & Linear & Linear & LayerNorm & LayerNorm \\
 & \multirow{2}{*}{Sigmoid} & \multirow{2}{*}{Sigmoid} & F.F. & F.F. \\
 &  &  & RC3b\textleftarrow & RC3b\textleftarrow \\
 \cmidrule(lr){2-2} \cmidrule(lr){3-3}   \cmidrule(lr){4-4}
\cmidrule(lr){5-5}
\multirow{8}{*}{4} & \multirow{15}{*}{N/A} & \multirow{15}{*}{N/A} & RC4a\textrightarrow & RC4a\textrightarrow \\
 &  &  & LayerNorm & LayerNorm \\
 &  &  & MSA(16,64) & MSA(4,256) \\
 &  &  & RC4a\textleftarrow & RC4a\textleftarrow \\
 &  &  & RC4b\textrightarrow & RC4b\textrightarrow \\
 &  &  & LayerNorm & LayerNorm \\
 &  &  & F.F. & F.F. \\
 &  &  & RC4b\textleftarrow & RC4b\textleftarrow \\
\cmidrule(lr){4-4}
\cmidrule(lr){5-5}
\multirow{7}{*}{5} &  &  & \verb|[token]|\textrightarrow & \verb|[token]|\textrightarrow  \\
 &  &  & LayerNorm & LayerNorm \\
 &  &  & Linear & Linear \\
 &  &  & GELU & GELU \\
 &  &  & Droput(0.5) & Droput(0.5) \\
 &  &  & Linear & Linear \\
 &  &  & Sigmoid & Sigmoid
 \\
 \bottomrule
 \end{tabularx}
\end{table}

%% file: tables/supp/net_architecture/lstm-rnn-arch.tex
\begin{table}[]
\caption{The architectures of LSTM-\textalpha, LSTM-\textbeta, RNN-\textalpha, and RNN-\textbeta. We post-append the hidden state units to each \protect\say{RNN Cell} and \protect\say{LSTM Cell}.
}
\centering
\label{supp_tab:lstm_rnn_arch}
\footnotesize
\begin{tabularx}{1.0\textwidth}{ijjjj}
\toprule
 Layer & LSTM-\textalpha & LSTM-\textbeta & RNN-\textalpha & RNN-\textbeta \\
\cmidrule(lr){1-1} \cmidrule(lr){2-2} \cmidrule(lr){3-3} \cmidrule(lr){4-4} \cmidrule(lr){5-5}
1 & LSTM Cell (256) & LSTM Cell (128) & RNN Cell (256) & RNN Cell (128) \\
\cmidrule(lr){2-2} \cmidrule(lr){3-3} \cmidrule(lr){4-4} \cmidrule(lr){5-5}
\multirow{2}{*}{2} & Linear & Linear & Linear & Linear \\
 & Sigmoid & Sigmoid & Sigmoid & Sigmoid
 \\
 \bottomrule
 \end{tabularx}
\end{table}

%% file: tables/supp/train_setting.tex
\begin{table}[h!]
\vspace{-2em}
\caption{The training setting for 11 DNNs across three datasets. We use BCE loss and Adam optimiser for all experiments. \protect\say{Init. LR} refers to the initial learning rate for Adam optimiser, and \protect\say{min(64, $\abs{\mathcal{C}}$)} stands for using the smaller of 64 and $\abs{\mathcal{C}}$ (size of condensed set) as the batch size.
}
\centering
\label{supp_tab:train_setting}
\footnotesize
\begin{tabularx}{0.9\textwidth}{jyjyy}
\toprule
\multirow{2}{*}{Dataset} & \multirow{2}{*}{Train data} & \multicolumn{3}{c}{Hyper-parameters} \\
 &  & Batch Size & Init. LR & Epochs \\
 \cmidrule(lr){1-1}  \cmidrule(lr){2-2} \cmidrule(lr){3-5} 
\multirow{2}{*}{PhysioNet-2012} & Original & 64 & \multirow{2}{*}{\numprint{5e-4}} & \multirow{2}{*}{30} \\
 & Condensed & min(64, $\abs{\mathcal{C}}$) &  &  \\
 \cmidrule(lr){2-2} \cmidrule(lr){3-5} 
\multirow{2}{*}{MIMIC-III} & Original & \multirow{2}{*}{64} & \multirow{2}{*}{\numprint{5e-5}} & \multirow{2}{*}{20} \\
 & Condensed &  &  &  \\
 \cmidrule(lr){2-2} \cmidrule(lr){3-5} 
\multirow{2}{*}{Coswara} & Original & 64 & \multirow{2}{*}{\numprint{1e-4}} & \multirow{2}{*}{30} \\
 & Condensed & min(64, $\abs{\mathcal{C}}$) &  & 
 \\
 \bottomrule
 \end{tabularx}
\end{table}

%% file: tables/supp/results/pre_dnn_res.tex
\begin{table}[h!]
\vspace{-2em}
\centering
\caption{The AUC of DNNs trained on original and condensed sets across all 3 datasets and 11 networks. We repeat each training session for five times and report the average AUC with STD on the test set.
\protect\say{AVG} refers to the average AUC of all sessions.
}
\label{supp_tab:per_dnn_res}
\footnotesize
\begin{tabularx}{1.0\textwidth}{yydddd}
\toprule
\multirow{2}{*}{\textbf{Dataset}} & \multirow{2}{*}{\shortstack{\textbf{Train data}\\(samples)}} & \multicolumn{4}{c}{\textbf{Test AUC} (SD)} \\
 &  & TCN-\textalpha & TCN-\textbeta & TCN-\textgamma & ViT-\textalpha \\
 \cmidrule(lr){1-2} \cmidrule(lr){3-6}
\multirow{4}{*}{\shortstack{PhysioNet\\-2012}} & Ori.(5120) & 0.854 (0.005) & 0.858 (0.005) & 0.852 (0.006) & 0.854 (0.003) \\
 & Con.(80) & 0.806 (0.003) & 0.800 (0.003) & 0.804 (0.006) & 0.798 (0.008) \\
 & Con.(40) & 0.813 (0.003) & 0.804 (0.005) & 0.808 (0.005) & 0.788 (0.007) \\
 & Con.(20) & 0.808 (0.006) & 0.807 (0.002) & 0.808 (0.002) & 0.806 (0.011) \\
 \cmidrule(lr){3-6}
\multirow{4}{*}{MIMIC-III} & Ori.(14698) & 0.843 (0.004) & 0.833 (0.002) & 0.845 (0.001) & 0.835 (0.002) \\
 & Con.(1200) & 0.756 (0.001) & 0.750 (0.006) & 0.756 (0.002) & 0.743 (0.011) \\
 & Con.(800) & 0.748 (0.002) & 0.736 (0.009) & 0.747 (0.000) & 0.744 (0.007) \\
 & Con.(400) & 0.740 (0.006) & 0.704 (0.021) & 0.742 (0.007) & 0.740 (0.008) \\
 \cmidrule(lr){3-6}
\multirow{4}{*}{Coswara} & Ori.(987) & 0.729 (0.017) & 0.750 (0.007) & 0.740 (0.018) & 0.727 (0.019) \\
 & Con.(80) & 0.646 (0.010) & 0.635 (0.006) & 0.641 (0.002) & 0.637 (0.030) \\
 & Con.(40) & 0.645 (0.007) & 0.636 (0.008) & 0.650 (0.014) & 0.626 (0.021) \\
 & Con.(20) & 0.647 (0.006) & 0.633 (0.005) & 0.651 (0.012) & 0.590 (0.022) \\
\cmidrule(lr){3-6}
\multicolumn{1}{l}{} & \multicolumn{1}{l}{} & ViT-\textbeta & TRSF-\textalpha & TRSF-\textbeta & LSTM-\textalpha \\
 \cmidrule(lr){1-2} \cmidrule(lr){3-6}
\multirow{4}{*}{\shortstack{PhysioNet\\-2012}} & Ori.(5120) & 0.851 (0.005) & 0.853 (0.003) & 0.843 (0.016) & 0.869 (0.003) \\
 & Con.(80) & 0.798 (0.010) & 0.781 (0.006) & 0.788 (0.005) & 0.815 (0.004) \\
 & Con.(40) & 0.797 (0.006) & 0.779 (0.007) & 0.784 (0.011) & 0.816 (0.007) \\
 & Con.(20) & 0.794 (0.014) & 0.789 (0.007) & 0.787 (0.004) & 0.804 (0.005) \\
 \cmidrule(lr){3-6}
\multirow{4}{*}{MIMIC-III} & Ori.(14698) & 0.830 (0.003) & 0.835 (0.001) & 0.832 (0.004) & 0.849 (0.004) \\
 & Con.(1200) & 0.745 (0.007) & 0.742 (0.005) & 0.743 (0.005) & 0.776 (0.003) \\
 & Con.(800) & 0.741 (0.006) & 0.734 (0.006) & 0.742 (0.006) & 0.777 (0.011) \\
 & Con.(400) & 0.737 (0.005) & 0.735 (0.009) & 0.737 (0.008) & 0.770 (0.010) \\
 \cmidrule(lr){3-6}
\multirow{4}{*}{Coswara} & Ori.(987) & 0.738 (0.020) & 0.732 (0.017) & 0.741 (0.020) & 0.739 (0.005) \\
 & Con.(80) & 0.634 (0.009) & 0.642 (0.019) & 0.629 (0.017) & 0.654 (0.006) \\
 & Con.(40) & 0.614 (0.017) & 0.630 (0.034) & 0.643 (0.026) & 0.654 (0.007) \\
 & Con.(20) & 0.618 (0.029) & 0.603 (0.024) & 0.602 (0.028) & 0.663 (0.013) \\
\cmidrule(lr){3-6}
\multicolumn{1}{l}{} & \multicolumn{1}{l}{} & AVG & LSTM-\textbeta & RNN-\textalpha & RNN2-\textbeta \\
   \cmidrule(lr){1-2} \cmidrule(lr){3-6}
\multirow{4}{*}{\shortstack{PhysioNet\\-2012}} & Ori.(5120) & 0.858 (0.011) & 0.873 (0.003) & 0.864 (0.009) & 0.872 (0.005) \\
 & Con.(80) & 0.804 (0.014) & 0.817 (0.011) & 0.820 (0.004) & 0.820 (0.006) \\
 & Con.(40) & 0.804 (0.016) & 0.813 (0.012) & 0.822 (0.008) & 0.815 (0.011) \\
 & Con.(20) & 0.803 (0.012) & 0.816 (0.005) & 0.811 (0.013) & 0.804 (0.015) \\
  \cmidrule(lr){3-6}
\multirow{4}{*}{MIMIC-III} & Ori.(14698) & 0.840 (0.007) & 0.844 (0.002) & 0.847 (0.001) & 0.845 (0.002) \\
 & Con.(1200) & 0.756 (0.014) & 0.774 (0.007) & 0.768 (0.004) & 0.761 (0.013) \\
 & Con.(800) & 0.750 (0.014) & 0.761 (0.006) & 0.757 (0.006) & 0.761 (0.012) \\
 & Con.(400) & 0.741 (0.019) & 0.749 (0.016) & 0.760 (0.007) & 0.738 (0.012) \\
  \cmidrule(lr){3-6}
\multirow{4}{*}{Coswara} & Ori.(987) & 0.737 (0.017) & 0.735 (0.007) & 0.753 (0.005) & 0.721 (0.014) \\
 & Con.(80) & 0.642 (0.016) & 0.651 (0.011) & 0.648 (0.005) & 0.640 (0.013) \\
 & Con.(40) & 0.641 (0.021) & 0.652 (0.012) & 0.658 (0.009) & 0.645 (0.008) \\
 & Con.(20) & 0.632 (0.030) & 0.653 (0.010) & 0.648 (0.009) & 0.647 (0.011)
 \\
 \bottomrule
 \end{tabularx}
\end{table}

%% file: tables/supp/results/time_dim_effect.tex
\begin{table}[h!]
\vspace{-2em}
\caption{The effects of different temporal dimensions of condensed samples across 3 datasets. $\abs{\mathcal{C}}$ refers to the number of samples in the condensed set, while \protect\say{$T$ (Ori.)}  and \protect\say{$T$ (Con.)} are abbreviations of the time dimension in the original and condensed samples, respectively.
Reducing $T^{\ast}$ can further compress the data volumes without significantly affecting the test AUCs, reflecting high flexibility in size control using DC.
}
\centering
\label{supp_tab:time_dim_effects}
\footnotesize
\begin{tabularx}{1.0\textwidth}{dggyyj}
\toprule
\multirow{2}{*}{\textbf{Dataset}} & \multirow{2}{*}{$\abs{\mathcal{C}}$} & \multicolumn{2}{c}{\textbf{Temporal Dimension}} & \multirow{2}{*}{\textbf{Sizes} (MBs)} & \multirow{2}{*}{\textbf{Test AUC} (SD)} \\
 &  & $T$ (Ori.) & $T^{\ast}$ (Con.) &  &  \\
  \cmidrule(lr){1-2} \cmidrule(lr){3-4}  \cmidrule(lr){5-5} \cmidrule(lr){6-6}
\multirow{4}{*}{PhysioNet-2012} & \multirow{4}{*}{80} & \multirow{4}{*}{48} & 48 & 0.69 & 0.804 (0.014) \\
 &  &  & 36 & 0.52 & 0.799 (0.015) \\
 &  &  & 24 & 0.34 & 0.801 (0.016) \\
 &  &  & 12 & 0.17 & 0.801 (0.016) \\
 \cmidrule(lr){4-4} \cmidrule(lr){5-5} \cmidrule(lr){6-6}
\multirow{4}{*}{MIMIC-III} & \multirow{4}{*}{1200} & \multirow{4}{*}{48} & 48 & 13.18 & 0.756 (0.014) \\
 &  &  & 36 & 9.89 & 0.745 (0.017) \\
 &  &  & 24 & 6.59 & 0.747 (0.017) \\
 &  &  & 12 & 3.30 & 0.746 (0.018) \\
  \cmidrule(lr){4-4} \cmidrule(lr){5-5} \cmidrule(lr){6-6}
\multirow{4}{*}{Coswara} & \multirow{4}{*}{80} & \multirow{4}{*}{96} & 96 & 1.88 & 0.642 (0.016) \\
 &  &  & 64 & 1.25 & 0.639 (0.021) \\
 &  &  & 32 & 0.63 & 0.631 (0.034) \\
 &  &  & 16 & 0.31 & 0.632 (0.030)
 \\
 \bottomrule
 \end{tabularx}
\end{table}

%% file: tables/supp/time_costs.tex
\begin{table}[h!]
\vspace{-2em}
\caption{Training time required to learn condensed sets of different sizes across 3 datasets. 
We measure the duration (minutes) of each DC learning session of \numprint{24000} iterations on an NVIDIA GeForce RTX 3060 graphic card.
All condensed datasets can be learned within approximately 1 hour.
}
\centering
\label{supp_tab:dc_time_cost}
\begin{tabularx}{0.87\textwidth}{>{\raggedright\arraybackslash}uuu}
\toprule
\textbf{Dataset} (samples) & \textbf{Condensed size} & \textbf{DC Time} (min) \\
 \cmidrule(lr){1-1}  \cmidrule(lr){2-2}  \cmidrule(l){3-3} 
\multirow{3}{*}{PhysioNet-2012 (5120)} & 80 & 18.3 \\
 & 40 & 16.9 \\
 & 20 & 16.2 \\
 \cmidrule(lr){2-2}  \cmidrule(l){3-3} 
\multirow{3}{*}{MIMIC-III (14698)} & 1200 & 61.8 \\
 & 800 & 46.0 \\
 & 400 & 30.6 \\
 \cmidrule(lr){2-2}  \cmidrule(l){3-3} 
\multirow{3}{*}{Coswara (987)} & 80 & 31.1 \\
 & 40 & 28.2 \\
 & 20 & 27.4
 \\
 \bottomrule
 \end{tabularx}
\end{table}

%% file: main.bbl

\begin{thebibliography}{31}
\ifx \bisbn   \undefined \def \bisbn  #1{ISBN #1}\fi
\ifx \binits  \undefined \def \binits#1{#1}\fi
\ifx \bauthor  \undefined \def \bauthor#1{#1}\fi
\ifx \batitle  \undefined \def \batitle#1{#1}\fi
\ifx \bjtitle  \undefined \def \bjtitle#1{#1}\fi
\ifx \bvolume  \undefined \def \bvolume#1{\textbf{#1}}\fi
\ifx \byear  \undefined \def \byear#1{#1}\fi
\ifx \bissue  \undefined \def \bissue#1{#1}\fi
\ifx \bfpage  \undefined \def \bfpage#1{#1}\fi
\ifx \blpage  \undefined \def \blpage #1{#1}\fi
\ifx \burl  \undefined \def \burl#1{\textsf{#1}}\fi
\ifx \doiurl  \undefined \def \doiurl#1{\url{https://doi.org/#1}}\fi
\ifx \betal  \undefined \def \betal{\textit{et al.}}\fi
\ifx \binstitute  \undefined \def \binstitute#1{#1}\fi
\ifx \binstitutionaled  \undefined \def \binstitutionaled#1{#1}\fi
\ifx \bctitle  \undefined \def \bctitle#1{#1}\fi
\ifx \beditor  \undefined \def \beditor#1{#1}\fi
\ifx \bpublisher  \undefined \def \bpublisher#1{#1}\fi
\ifx \bbtitle  \undefined \def \bbtitle#1{#1}\fi
\ifx \bedition  \undefined \def \bedition#1{#1}\fi
\ifx \bseriesno  \undefined \def \bseriesno#1{#1}\fi
\ifx \blocation  \undefined \def \blocation#1{#1}\fi
\ifx \bsertitle  \undefined \def \bsertitle#1{#1}\fi
\ifx \bsnm \undefined \def \bsnm#1{#1}\fi
\ifx \bsuffix \undefined \def \bsuffix#1{#1}\fi
\ifx \bparticle \undefined \def \bparticle#1{#1}\fi
\ifx \barticle \undefined \def \barticle#1{#1}\fi
\bibcommenthead
\ifx \bconfdate \undefined \def \bconfdate #1{#1}\fi
\ifx \botherref \undefined \def \botherref #1{#1}\fi
\ifx \url \undefined \def \url#1{\textsf{#1}}\fi
\ifx \bchapter \undefined \def \bchapter#1{#1}\fi
\ifx \bbook \undefined \def \bbook#1{#1}\fi
\ifx \bcomment \undefined \def \bcomment#1{#1}\fi
\ifx \oauthor \undefined \def \oauthor#1{#1}\fi
\ifx \citeauthoryear \undefined \def \citeauthoryear#1{#1}\fi
\ifx \endbibitem  \undefined \def \endbibitem {}\fi
\ifx \bconflocation  \undefined \def \bconflocation#1{#1}\fi
\ifx \arxivurl  \undefined \def \arxivurl#1{\textsf{#1}}\fi
\csname PreBibitemsHook\endcsname

\bibitem{topol2019high}
\begin{barticle}
\bauthor{\bsnm{Topol}, \binits{E.J.}}:
\batitle{High-performance medicine: the convergence of human and artificial intelligence}.
\bjtitle{Nature medicine}
\bvolume{25}(\bissue{1}),
\bfpage{44}--\blpage{56}
(\byear{2019})
\end{barticle}
\endbibitem

\bibitem{zhang2022shifting}
\begin{botherref}
\oauthor{\bsnm{Zhang}, \binits{A.}},
\oauthor{\bsnm{Xing}, \binits{L.}},
\oauthor{\bsnm{Zou}, \binits{J.}},
\oauthor{\bsnm{Wu}, \binits{J.C.}}:
Shifting machine learning for healthcare from development to deployment and from models to data.
Nature Biomedical Engineering,
1--16
(2022)
\end{botherref}
\endbibitem

\bibitem{ReportsAndData2021a}
\begin{botherref}
Global Artificial Intelligence In Healthcare Market Size To Worth USD 280.77 Billion By 2032 - CAGR of 30.6\%.
SPHERICAL INSIGHTS LLP.
Accessed on Aug 15th, 2023
(2023).
\url{https://www.globenewswire.com/en/news-release/2023/07/04/2698952/0/en/Global-Artificial-Intelligence-In-Healthcare-Market-Size-To-Worth-USD-280-77-Billion-By-2032-CAGR-of-30-6.html}
\end{botherref}
\endbibitem

\bibitem{Medicine2018}
\begin{botherref}
The Democratization of Health Care.
Stanford Medicine.
Accessed on June 15th, 2023
(2018).
\url{https://med.stanford.edu/content/dam/sm/school/documents/Health-Trends-Report/Stanford-Medicine-Health-Trends-Report-2018.pdf}
\end{botherref}
\endbibitem

\bibitem{wang2022reinforcing}
\begin{barticle}
\bauthor{\bsnm{Wang}, \binits{Y.}},
\bauthor{\bsnm{Blobel}, \binits{B.}},
\bauthor{\bsnm{Yang}, \binits{B.}}:
\batitle{Reinforcing health data sharing through data democratization}.
\bjtitle{Journal of Personalized Medicine}
\bvolume{12}(\bissue{9}),
\bfpage{1380}
(\byear{2022})
\end{barticle}
\endbibitem

\bibitem{lewis2020data}
\begin{bchapter}
\bauthor{\bsnm{Lewis}, \binits{K.}},
\bauthor{\bsnm{Pham}, \binits{C.}},
\bauthor{\bsnm{Batarseh}, \binits{F.A.}}:
\bctitle{Data openness and democratization in healthcare: An evaluation of hospital ranking methods}.
In: \bbtitle{Data Democracy},
pp. \bfpage{109}--\blpage{126}.
\bpublisher{Elsevier}, \blocation{???}
(\byear{2020})
\end{bchapter}
\endbibitem

\bibitem{rajkomar2018scalable}
\begin{barticle}
\bauthor{\bsnm{Rajkomar}, \binits{A.}},
\bauthor{\bsnm{Oren}, \binits{E.}},
\bauthor{\bsnm{Chen}, \binits{K.}},
\bauthor{\bsnm{Dai}, \binits{A.M.}},
\bauthor{\bsnm{Hajaj}, \binits{N.}},
\bauthor{\bsnm{Hardt}, \binits{M.}},
\bauthor{\bsnm{Liu}, \binits{P.J.}},
\bauthor{\bsnm{Liu}, \binits{X.}},
\bauthor{\bsnm{Marcus}, \binits{J.}},
\bauthor{\bsnm{Sun}, \binits{M.}}, \betal:
\batitle{Scalable and accurate deep learning with electronic health records}.
\bjtitle{NPJ digital medicine}
\bvolume{1}(\bissue{1}),
\bfpage{1}--\blpage{10}
(\byear{2018})
\end{barticle}
\endbibitem

\bibitem{el2013anonymizing}
\begin{bbook}
\bauthor{\bsnm{El~Emam}, \binits{K.}},
\bauthor{\bsnm{Arbuckle}, \binits{L.}}:
\bbtitle{Anonymizing Health Data: Case Studies and Methods to Get You Started},
\bedition{1st} edn.
\bpublisher{O'Reilly Media, Inc.},
\blocation{Sebastopol}
(\byear{2013})
\end{bbook}
\endbibitem

\bibitem{narayanan2010myths}
\begin{barticle}
\bauthor{\bsnm{Narayanan}, \binits{A.}},
\bauthor{\bsnm{Shmatikov}, \binits{V.}}:
\batitle{Myths and fallacies of" personally identifiable information"}.
\bjtitle{Communications of the ACM}
\bvolume{53}(\bissue{6}),
\bfpage{24}--\blpage{26}
(\byear{2010})
\end{barticle}
\endbibitem

\bibitem{sweeney2015anonymizing}
\begin{botherref}
\oauthor{\bsnm{Sweeney}, \binits{L.}},
\oauthor{\bsnm{Yoo}, \binits{J.S.}}:
De-anonymizing south korean resident registration numbers shared in prescription data.
Technology Science
(2015)
\end{botherref}
\endbibitem

\bibitem{de2015unique}
\begin{barticle}
\bauthor{\bsnm{De~Montjoye}, \binits{Y.-A.}},
\bauthor{\bsnm{Radaelli}, \binits{L.}},
\bauthor{\bsnm{Singh}, \binits{V.K.}},
\bauthor{\bsnm{Pentland}, \binits{A.S.}}:
\batitle{Unique in the shopping mall: On the reidentifiability of credit card metadata}.
\bjtitle{Science}
\bvolume{347}(\bissue{6221}),
\bfpage{536}--\blpage{539}
(\byear{2015})
\end{barticle}
\endbibitem

\bibitem{pandey2022transformational}
\begin{barticle}
\bauthor{\bsnm{Pandey}, \binits{M.}},
\bauthor{\bsnm{Fernandez}, \binits{M.}},
\bauthor{\bsnm{Gentile}, \binits{F.}},
\bauthor{\bsnm{Isayev}, \binits{O.}},
\bauthor{\bsnm{Tropsha}, \binits{A.}},
\bauthor{\bsnm{Stern}, \binits{A.C.}},
\bauthor{\bsnm{Cherkasov}, \binits{A.}}:
\batitle{The transformational role of gpu computing and deep learning in drug discovery}.
\bjtitle{Nature Machine Intelligence}
\bvolume{4}(\bissue{3}),
\bfpage{211}--\blpage{221}
(\byear{2022})
\end{barticle}
\endbibitem

\bibitem{zhao2021distributionmatching}
\begin{bchapter}
\bauthor{\bsnm{Zhao}, \binits{B.}},
\bauthor{\bsnm{Bilen}, \binits{H.}}:
\bctitle{Dataset condensation with distribution matching}.
In: \bbtitle{Proceedings of the IEEE/CVF Winter Conference on Applications of Computer Vision},
pp. \bfpage{6514}--\blpage{6523}
(\byear{2023})
\end{bchapter}
\endbibitem

\bibitem{silva2012predicting}
\begin{bchapter}
\bauthor{\bsnm{Silva}, \binits{I.}},
\bauthor{\bsnm{Moody}, \binits{G.}},
\bauthor{\bsnm{Scott}, \binits{D.J.}},
\bauthor{\bsnm{Celi}, \binits{L.A.}},
\bauthor{\bsnm{Mark}, \binits{R.G.}}:
\bctitle{Predicting in-hospital mortality of icu patients: The physionet/computing in cardiology challenge 2012}.
In: \bbtitle{2012 Computing in Cardiology},
pp. \bfpage{245}--\blpage{248}
(\byear{2012}).
\bcomment{IEEE}
\end{bchapter}
\endbibitem

\bibitem{johnson2016mimic}
\begin{barticle}
\bauthor{\bsnm{Johnson}, \binits{A.E.}},
\bauthor{\bsnm{Pollard}, \binits{T.J.}},
\bauthor{\bsnm{Shen}, \binits{L.}},
\bauthor{\bsnm{Lehman}, \binits{L.-w.H.}},
\bauthor{\bsnm{Feng}, \binits{M.}},
\bauthor{\bsnm{Ghassemi}, \binits{M.}},
\bauthor{\bsnm{Moody}, \binits{B.}},
\bauthor{\bsnm{Szolovits}, \binits{P.}},
\bauthor{\bsnm{Anthony~Celi}, \binits{L.}},
\bauthor{\bsnm{Mark}, \binits{R.G.}}:
\batitle{Mimic-iii, a freely accessible critical care database}.
\bjtitle{Scientific data}
\bvolume{3}(\bissue{1}),
\bfpage{1}--\blpage{9}
(\byear{2016})
\end{barticle}
\endbibitem

\bibitem{johnson2016mimic-physionet}
\begin{barticle}
\bauthor{\bsnm{Johnson}, \binits{A.}},
\bauthor{\bsnm{Pollard}, \binits{T.}},
\bauthor{\bsnm{Mark}, \binits{R.}}:
\batitle{Mimic-iii clinical database (version 1.4)}.
\bjtitle{PhysioNet}
\bvolume{10},
\bfpage{2}--\blpage{26}
(\byear{2016})
\end{barticle}
\endbibitem

\bibitem{sharma2020coswara}
\begin{bchapter}
\bauthor{\bsnm{Sharma}, \binits{N.}},
\bauthor{\bsnm{Krishnan}, \binits{P.}},
\bauthor{\bsnm{Kumar}, \binits{R.}},
\bauthor{\bsnm{Ramoji}, \binits{S.}},
\bauthor{\bsnm{Chetupalli}, \binits{S.}},
\bauthor{\bsnm{Nirmala}, \binits{R.}},
\bauthor{\bsnm{Kumar~Ghosh}, \binits{P.}},
\bauthor{\bsnm{Ganapathy}, \binits{S.}}:
\bctitle{Coswara-a database of breathing, cough, and voice sounds for covid-19 diagnosis}.
In: \bbtitle{Proceedings of the Annual Conference of the International Speech Communication Association, INTERSPEECH},
vol. \bseriesno{2020},
pp. \bfpage{4811}--\blpage{4815}
(\byear{2020}).
\bcomment{International Speech Communication Association}
\end{bchapter}
\endbibitem

\bibitem{martinez2020lipreading}
\begin{bchapter}
\bauthor{\bsnm{Martinez}, \binits{B.}},
\bauthor{\bsnm{Ma}, \binits{P.}},
\bauthor{\bsnm{Petridis}, \binits{S.}},
\bauthor{\bsnm{Pantic}, \binits{M.}}:
\bctitle{Lipreading using temporal convolutional networks}.
In: \bbtitle{ICASSP 2020-2020 IEEE International Conference on Acoustics, Speech and Signal Processing (ICASSP)},
pp. \bfpage{6319}--\blpage{6323}
(\byear{2020}).
\bcomment{IEEE}
\end{bchapter}
\endbibitem

\bibitem{dosovitskiy2021an}
\begin{bchapter}
\bauthor{\bsnm{Dosovitskiy}, \binits{A.}},
\bauthor{\bsnm{Beyer}, \binits{L.}},
\bauthor{\bsnm{Kolesnikov}, \binits{A.}},
\bauthor{\bsnm{Weissenborn}, \binits{D.}},
\bauthor{\bsnm{Zhai}, \binits{X.}},
\bauthor{\bsnm{Unterthiner}, \binits{T.}},
\bauthor{\bsnm{Dehghani}, \binits{M.}},
\bauthor{\bsnm{Minderer}, \binits{M.}},
\bauthor{\bsnm{Heigold}, \binits{G.}},
\bauthor{\bsnm{Gelly}, \binits{S.}},
\bauthor{\bsnm{Uszkoreit}, \binits{J.}},
\bauthor{\bsnm{Houlsby}, \binits{N.}}:
\bctitle{An image is worth 16x16 words: Transformers for image recognition at scale}.
In: \bbtitle{International Conference on Learning Representations}
(\byear{2021}).
\burl{https://openreview.net/forum?id=YicbFdNTTy}
\end{bchapter}
\endbibitem

\bibitem{vaswani2017attention}
\begin{botherref}
\oauthor{\bsnm{Vaswani}, \binits{A.}},
\oauthor{\bsnm{Shazeer}, \binits{N.}},
\oauthor{\bsnm{Parmar}, \binits{N.}},
\oauthor{\bsnm{Uszkoreit}, \binits{J.}},
\oauthor{\bsnm{Jones}, \binits{L.}},
\oauthor{\bsnm{Gomez}, \binits{A.N.}},
\oauthor{\bsnm{Kaiser}, \binits{{\L}.}},
\oauthor{\bsnm{Polosukhin}, \binits{I.}}:
Attention is all you need.
Advances in neural information processing systems
\textbf{30}
(2017)
\end{botherref}
\endbibitem

\bibitem{hochreiter1997long}
\begin{barticle}
\bauthor{\bsnm{Hochreiter}, \binits{S.}},
\bauthor{\bsnm{Schmidhuber}, \binits{J.}}:
\batitle{Long short-term memory}.
\bjtitle{Neural computation}
\bvolume{9}(\bissue{8}),
\bfpage{1735}--\blpage{1780}
(\byear{1997})
\end{barticle}
\endbibitem

\bibitem{hopfield1982neural}
\begin{barticle}
\bauthor{\bsnm{Hopfield}, \binits{J.J.}}:
\batitle{Neural networks and physical systems with emergent collective computational abilities.}
\bjtitle{Proceedings of the national academy of sciences}
\bvolume{79}(\bissue{8}),
\bfpage{2554}--\blpage{2558}
(\byear{1982})
\end{barticle}
\endbibitem

\bibitem{van2008visualizing}
\begin{botherref}
\oauthor{\bparticle{Van~der} \bsnm{Maaten}, \binits{L.}},
\oauthor{\bsnm{Hinton}, \binits{G.}}:
Visualizing data using t-sne.
Journal of machine learning research
\textbf{9}(11)
(2008)
\end{botherref}
\endbibitem

\bibitem{goodfellow2020generative}
\begin{barticle}
\bauthor{\bsnm{Goodfellow}, \binits{I.}},
\bauthor{\bsnm{Pouget-Abadie}, \binits{J.}},
\bauthor{\bsnm{Mirza}, \binits{M.}},
\bauthor{\bsnm{Xu}, \binits{B.}},
\bauthor{\bsnm{Warde-Farley}, \binits{D.}},
\bauthor{\bsnm{Ozair}, \binits{S.}},
\bauthor{\bsnm{Courville}, \binits{A.}},
\bauthor{\bsnm{Bengio}, \binits{Y.}}:
\batitle{Generative adversarial networks}.
\bjtitle{Communications of the ACM}
\bvolume{63}(\bissue{11}),
\bfpage{139}--\blpage{144}
(\byear{2020})
\end{barticle}
\endbibitem

\bibitem{xu2019modeling}
\begin{botherref}
\oauthor{\bsnm{Xu}, \binits{L.}},
\oauthor{\bsnm{Skoularidou}, \binits{M.}},
\oauthor{\bsnm{Cuesta-Infante}, \binits{A.}},
\oauthor{\bsnm{Veeramachaneni}, \binits{K.}}:
Modeling tabular data using conditional gan.
Advances in Neural Information Processing Systems
\textbf{32}
(2019)
\end{botherref}
\endbibitem

\bibitem{stadler2022synthetic}
\begin{bchapter}
\bauthor{\bsnm{Stadler}, \binits{T.}},
\bauthor{\bsnm{Oprisanu}, \binits{B.}},
\bauthor{\bsnm{Troncoso}, \binits{C.}}:
\bctitle{Synthetic data--anonymisation groundhog day}.
In: \bbtitle{31st USENIX Security Symposium (USENIX Security 22)},
pp. \bfpage{1451}--\blpage{1468}
(\byear{2022})
\end{bchapter}
\endbibitem

\bibitem{froelicher2021truly}
\begin{barticle}
\bauthor{\bsnm{Froelicher}, \binits{D.}},
\bauthor{\bsnm{Troncoso-Pastoriza}, \binits{J.R.}},
\bauthor{\bsnm{Raisaro}, \binits{J.L.}},
\bauthor{\bsnm{Cuendet}, \binits{M.A.}},
\bauthor{\bsnm{Sousa}, \binits{J.S.}},
\bauthor{\bsnm{Cho}, \binits{H.}},
\bauthor{\bsnm{Berger}, \binits{B.}},
\bauthor{\bsnm{Fellay}, \binits{J.}},
\bauthor{\bsnm{Hubaux}, \binits{J.-P.}}:
\batitle{Truly privacy-preserving federated analytics for precision medicine with multiparty homomorphic encryption}.
\bjtitle{Nature communications}
\bvolume{12}(\bissue{1}),
\bfpage{1}--\blpage{10}
(\byear{2021})
\end{barticle}
\endbibitem

\bibitem{gretton2012kernel}
\begin{barticle}
\bauthor{\bsnm{Gretton}, \binits{A.}},
\bauthor{\bsnm{Borgwardt}, \binits{K.M.}},
\bauthor{\bsnm{Rasch}, \binits{M.J.}},
\bauthor{\bsnm{Sch{\"o}lkopf}, \binits{B.}},
\bauthor{\bsnm{Smola}, \binits{A.}}:
\batitle{A kernel two-sample test}.
\bjtitle{The Journal of Machine Learning Research}
\bvolume{13}(\bissue{1}),
\bfpage{723}--\blpage{773}
(\byear{2012})
\end{barticle}
\endbibitem

\bibitem{saxe2011random}
\begin{bchapter}
\bauthor{\bsnm{Saxe}, \binits{A.M.}},
\bauthor{\bsnm{Koh}, \binits{P.W.}},
\bauthor{\bsnm{Chen}, \binits{Z.}},
\bauthor{\bsnm{Bhand}, \binits{M.}},
\bauthor{\bsnm{Suresh}, \binits{B.}},
\bauthor{\bsnm{Ng}, \binits{A.Y.}}:
\bctitle{On random weights and unsupervised feature learning}.
In: \bbtitle{Icml}
(\byear{2011})
\end{bchapter}
\endbibitem

\bibitem{giryes2016deep}
\begin{barticle}
\bauthor{\bsnm{Giryes}, \binits{R.}},
\bauthor{\bsnm{Sapiro}, \binits{G.}},
\bauthor{\bsnm{Bronstein}, \binits{A.M.}}:
\batitle{Deep neural networks with random gaussian weights: A universal classification strategy?}
\bjtitle{IEEE Transactions on Signal Processing}
\bvolume{64}(\bissue{13}),
\bfpage{3444}--\blpage{3457}
(\byear{2016})
\end{barticle}
\endbibitem

\bibitem{kingma2014adam}
\begin{botherref}
\oauthor{\bsnm{Kingma}, \binits{D.P.}},
\oauthor{\bsnm{Ba}, \binits{J.}}:
Adam: A method for stochastic optimization.
arXiv preprint arXiv:1412.6980
(2014)
\end{botherref}
\endbibitem

\end{thebibliography}



\begin{thebibliography}{16}
\ifx \bisbn   \undefined \def \bisbn  #1{ISBN #1}\fi
\ifx \binits  \undefined \def \binits#1{#1}\fi
\ifx \bauthor  \undefined \def \bauthor#1{#1}\fi
\ifx \batitle  \undefined \def \batitle#1{#1}\fi
\ifx \bjtitle  \undefined \def \bjtitle#1{#1}\fi
\ifx \bvolume  \undefined \def \bvolume#1{\textbf{#1}}\fi
\ifx \byear  \undefined \def \byear#1{#1}\fi
\ifx \bissue  \undefined \def \bissue#1{#1}\fi
\ifx \bfpage  \undefined \def \bfpage#1{#1}\fi
\ifx \blpage  \undefined \def \blpage #1{#1}\fi
\ifx \burl  \undefined \def \burl#1{\textsf{#1}}\fi
\ifx \doiurl  \undefined \def \doiurl#1{\url{https://doi.org/#1}}\fi
\ifx \betal  \undefined \def \betal{\textit{et al.}}\fi
\ifx \binstitute  \undefined \def \binstitute#1{#1}\fi
\ifx \binstitutionaled  \undefined \def \binstitutionaled#1{#1}\fi
\ifx \bctitle  \undefined \def \bctitle#1{#1}\fi
\ifx \beditor  \undefined \def \beditor#1{#1}\fi
\ifx \bpublisher  \undefined \def \bpublisher#1{#1}\fi
\ifx \bbtitle  \undefined \def \bbtitle#1{#1}\fi
\ifx \bedition  \undefined \def \bedition#1{#1}\fi
\ifx \bseriesno  \undefined \def \bseriesno#1{#1}\fi
\ifx \blocation  \undefined \def \blocation#1{#1}\fi
\ifx \bsertitle  \undefined \def \bsertitle#1{#1}\fi
\ifx \bsnm \undefined \def \bsnm#1{#1}\fi
\ifx \bsuffix \undefined \def \bsuffix#1{#1}\fi
\ifx \bparticle \undefined \def \bparticle#1{#1}\fi
\ifx \barticle \undefined \def \barticle#1{#1}\fi
\bibcommenthead
\ifx \bconfdate \undefined \def \bconfdate #1{#1}\fi
\ifx \botherref \undefined \def \botherref #1{#1}\fi
\ifx \url \undefined \def \url#1{\textsf{#1}}\fi
\ifx \bchapter \undefined \def \bchapter#1{#1}\fi
\ifx \bbook \undefined \def \bbook#1{#1}\fi
\ifx \bcomment \undefined \def \bcomment#1{#1}\fi
\ifx \oauthor \undefined \def \oauthor#1{#1}\fi
\ifx \citeauthoryear \undefined \def \citeauthoryear#1{#1}\fi
\ifx \endbibitem  \undefined \def \endbibitem {}\fi
\ifx \bconflocation  \undefined \def \bconflocation#1{#1}\fi
\ifx \arxivurl  \undefined \def \arxivurl#1{\textsf{#1}}\fi
\csname PreBibitemsHook\endcsname

\bibitem{harutyunyan2019multitask}
\begin{barticle}
\bauthor{\bsnm{Harutyunyan}, \binits{H.}},
\bauthor{\bsnm{Khachatrian}, \binits{H.}},
\bauthor{\bsnm{Kale}, \binits{D.C.}},
\bauthor{\bsnm{Ver~Steeg}, \binits{G.}},
\bauthor{\bsnm{Galstyan}, \binits{A.}}:
\batitle{Multitask learning and benchmarking with clinical time series data}.
\bjtitle{Scientific data}
\bvolume{6}(\bissue{1}),
\bfpage{1}--\blpage{18}
(\byear{2019})
\end{barticle}
\endbibitem

\bibitem{silva2012predicting}
\begin{bchapter}
\bauthor{\bsnm{Silva}, \binits{I.}},
\bauthor{\bsnm{Moody}, \binits{G.}},
\bauthor{\bsnm{Scott}, \binits{D.J.}},
\bauthor{\bsnm{Celi}, \binits{L.A.}},
\bauthor{\bsnm{Mark}, \binits{R.G.}}:
\bctitle{Predicting in-hospital mortality of icu patients: The physionet/computing in cardiology challenge 2012}.
In: \bbtitle{2012 Computing in Cardiology},
pp. \bfpage{245}--\blpage{248}
(\byear{2012}).
\bcomment{IEEE}
\end{bchapter}
\endbibitem

\bibitem{le1984simplified}
\begin{barticle}
\bauthor{\bsnm{Le~Gall}, \binits{J.-R.}},
\bauthor{\bsnm{Loirat}, \binits{P.}},
\bauthor{\bsnm{Alperovitch}, \binits{A.}},
\bauthor{\bsnm{Glaser}, \binits{P.}},
\bauthor{\bsnm{Granthil}, \binits{C.}},
\bauthor{\bsnm{Mathieu}, \binits{D.}},
\bauthor{\bsnm{Mercier}, \binits{P.}},
\bauthor{\bsnm{Thomas}, \binits{R.}},
\bauthor{\bsnm{Villers}, \binits{D.}}:
\batitle{A simplified acute physiology score for icu patients.}
\bjtitle{Critical care medicine}
\bvolume{12}(\bissue{11}),
\bfpage{975}--\blpage{977}
(\byear{1984})
\end{barticle}
\endbibitem

\bibitem{ferreira2001serial}
\begin{barticle}
\bauthor{\bsnm{Ferreira}, \binits{F.L.}},
\bauthor{\bsnm{Bota}, \binits{D.P.}},
\bauthor{\bsnm{Bross}, \binits{A.}},
\bauthor{\bsnm{M{\'e}lot}, \binits{C.}},
\bauthor{\bsnm{Vincent}, \binits{J.-L.}}:
\batitle{Serial evaluation of the sofa score to predict outcome in critically ill patients}.
\bjtitle{Jama}
\bvolume{286}(\bissue{14}),
\bfpage{1754}--\blpage{1758}
(\byear{2001})
\end{barticle}
\endbibitem

\bibitem{johnson2016mimic}
\begin{barticle}
\bauthor{\bsnm{Johnson}, \binits{A.E.}},
\bauthor{\bsnm{Pollard}, \binits{T.J.}},
\bauthor{\bsnm{Shen}, \binits{L.}},
\bauthor{\bsnm{Lehman}, \binits{L.-w.H.}},
\bauthor{\bsnm{Feng}, \binits{M.}},
\bauthor{\bsnm{Ghassemi}, \binits{M.}},
\bauthor{\bsnm{Moody}, \binits{B.}},
\bauthor{\bsnm{Szolovits}, \binits{P.}},
\bauthor{\bsnm{Anthony~Celi}, \binits{L.}},
\bauthor{\bsnm{Mark}, \binits{R.G.}}:
\batitle{Mimic-iii, a freely accessible critical care database}.
\bjtitle{Scientific data}
\bvolume{3}(\bissue{1}),
\bfpage{1}--\blpage{9}
(\byear{2016})
\end{barticle}
\endbibitem

\bibitem{johnson2016mimic-physionet}
\begin{barticle}
\bauthor{\bsnm{Johnson}, \binits{A.}},
\bauthor{\bsnm{Pollard}, \binits{T.}},
\bauthor{\bsnm{Mark}, \binits{R.}}:
\batitle{Mimic-iii clinical database (version 1.4)}.
\bjtitle{PhysioNet}
\bvolume{10},
\bfpage{2}--\blpage{26}
(\byear{2016})
\end{barticle}
\endbibitem

\bibitem{sharma2020coswara}
\begin{bchapter}
\bauthor{\bsnm{Sharma}, \binits{N.}},
\bauthor{\bsnm{Krishnan}, \binits{P.}},
\bauthor{\bsnm{Kumar}, \binits{R.}},
\bauthor{\bsnm{Ramoji}, \binits{S.}},
\bauthor{\bsnm{Chetupalli}, \binits{S.}},
\bauthor{\bsnm{Nirmala}, \binits{R.}},
\bauthor{\bsnm{Kumar~Ghosh}, \binits{P.}},
\bauthor{\bsnm{Ganapathy}, \binits{S.}}:
\bctitle{Coswara-a database of breathing, cough, and voice sounds for covid-19 diagnosis}.
In: \bbtitle{Proceedings of the Annual Conference of the International Speech Communication Association, INTERSPEECH},
vol. \bseriesno{2020},
pp. \bfpage{4811}--\blpage{4815}
(\byear{2020}).
\bcomment{International Speech Communication Association}
\end{bchapter}
\endbibitem

\bibitem{martinez2020lipreading}
\begin{bchapter}
\bauthor{\bsnm{Martinez}, \binits{B.}},
\bauthor{\bsnm{Ma}, \binits{P.}},
\bauthor{\bsnm{Petridis}, \binits{S.}},
\bauthor{\bsnm{Pantic}, \binits{M.}}:
\bctitle{Lipreading using temporal convolutional networks}.
In: \bbtitle{ICASSP 2020-2020 IEEE International Conference on Acoustics, Speech and Signal Processing (ICASSP)},
pp. \bfpage{6319}--\blpage{6323}
(\byear{2020}).
\bcomment{IEEE}
\end{bchapter}
\endbibitem

\bibitem{vaswani2017attention}
\begin{botherref}
\oauthor{\bsnm{Vaswani}, \binits{A.}},
\oauthor{\bsnm{Shazeer}, \binits{N.}},
\oauthor{\bsnm{Parmar}, \binits{N.}},
\oauthor{\bsnm{Uszkoreit}, \binits{J.}},
\oauthor{\bsnm{Jones}, \binits{L.}},
\oauthor{\bsnm{Gomez}, \binits{A.N.}},
\oauthor{\bsnm{Kaiser}, \binits{{\L}.}},
\oauthor{\bsnm{Polosukhin}, \binits{I.}}:
Attention is all you need.
Advances in neural information processing systems
\textbf{30}
(2017)
\end{botherref}
\endbibitem

\bibitem{dosovitskiy2021an}
\begin{bchapter}
\bauthor{\bsnm{Dosovitskiy}, \binits{A.}},
\bauthor{\bsnm{Beyer}, \binits{L.}},
\bauthor{\bsnm{Kolesnikov}, \binits{A.}},
\bauthor{\bsnm{Weissenborn}, \binits{D.}},
\bauthor{\bsnm{Zhai}, \binits{X.}},
\bauthor{\bsnm{Unterthiner}, \binits{T.}},
\bauthor{\bsnm{Dehghani}, \binits{M.}},
\bauthor{\bsnm{Minderer}, \binits{M.}},
\bauthor{\bsnm{Heigold}, \binits{G.}},
\bauthor{\bsnm{Gelly}, \binits{S.}},
\bauthor{\bsnm{Uszkoreit}, \binits{J.}},
\bauthor{\bsnm{Houlsby}, \binits{N.}}:
\bctitle{An image is worth 16x16 words: Transformers for image recognition at scale}.
In: \bbtitle{International Conference on Learning Representations}
(\byear{2021}).
\burl{https://openreview.net/forum?id=YicbFdNTTy}
\end{bchapter}
\endbibitem

\bibitem{hochreiter1997long}
\begin{barticle}
\bauthor{\bsnm{Hochreiter}, \binits{S.}},
\bauthor{\bsnm{Schmidhuber}, \binits{J.}}:
\batitle{Long short-term memory}.
\bjtitle{Neural computation}
\bvolume{9}(\bissue{8}),
\bfpage{1735}--\blpage{1780}
(\byear{1997})
\end{barticle}
\endbibitem

\bibitem{hopfield1982neural}
\begin{barticle}
\bauthor{\bsnm{Hopfield}, \binits{J.J.}}:
\batitle{Neural networks and physical systems with emergent collective computational abilities.}
\bjtitle{Proceedings of the national academy of sciences}
\bvolume{79}(\bissue{8}),
\bfpage{2554}--\blpage{2558}
(\byear{1982})
\end{barticle}
\endbibitem

\bibitem{he2015delving}
\begin{bchapter}
\bauthor{\bsnm{He}, \binits{K.}},
\bauthor{\bsnm{Zhang}, \binits{X.}},
\bauthor{\bsnm{Ren}, \binits{S.}},
\bauthor{\bsnm{Sun}, \binits{J.}}:
\bctitle{Delving deep into rectifiers: Surpassing human-level performance on imagenet classification}.
In: \bbtitle{Proceedings of the IEEE International Conference on Computer Vision},
pp. \bfpage{1026}--\blpage{1034}
(\byear{2015})
\end{bchapter}
\endbibitem

\bibitem{bai2018empirical}
\begin{botherref}
\oauthor{\bsnm{Bai}, \binits{S.}},
\oauthor{\bsnm{Kolter}, \binits{J.Z.}},
\oauthor{\bsnm{Koltun}, \binits{V.}}:
An empirical evaluation of generic convolutional and recurrent networks for sequence modeling.
arXiv preprint arXiv:1803.01271
(2018)
\end{botherref}
\endbibitem

\bibitem{hendrycks2016gaussian}
\begin{botherref}
\oauthor{\bsnm{Hendrycks}, \binits{D.}},
\oauthor{\bsnm{Gimpel}, \binits{K.}}:
Gaussian error linear units (gelus).
arXiv preprint arXiv:1606.08415
(2016)
\end{botherref}
\endbibitem

\bibitem{kingma2014adam}
\begin{botherref}
\oauthor{\bsnm{Kingma}, \binits{D.P.}},
\oauthor{\bsnm{Ba}, \binits{J.}}:
Adam: A method for stochastic optimization.
arXiv preprint arXiv:1412.6980
(2014)
\end{botherref}
\endbibitem

\end{thebibliography}
